\documentclass{article}
\usepackage[preprint]{tmlr}
\usepackage{times}
\usepackage{graphicx} 
\usepackage[a4paper,margin=1in,footskip=0.25in]{geometry}
\usepackage[utf8]{inputenc} 
\usepackage[T1]{fontenc}    
\usepackage{hyperref}       
\usepackage{url}            
\usepackage{booktabs}       
\usepackage{amsfonts}       
\usepackage{nicefrac}       
\usepackage{microtype}      
\usepackage{xcolor}         
\usepackage{bm}
\usepackage{tcolorbox}  
\usepackage[makeroom]{cancel}
\usepackage{natbib}
\usepackage{amsmath}
\usepackage{svg}
\usepackage{enumitem}
\usepackage{subcaption}
\usepackage{multirow}
\usepackage{makecell}

\newcommand{\pass}[1]{^{(#1)}}

\newcommand\cross[2]{#1 \otimes #2}

\newcommand{\embd}{\bm{e}}

\newcommand{\bWR}{\bm{W}^R}
\newcommand{\bWU}{\bm{W}^H}
\newcommand{\bWG}{\bm{W}^G}

\newcommand{\be}{\bm{e}}

\newcommand{\bu}{\bm{u}}

\newcommand{\R}{\mathbb{R}}

\newcommand{\bh}{\bm{h}}

\newcommand\E[2]{\operatornamewithlimits{\mathbb{E}}_{#1}\left[#2\right]}

\newcommand{\norm}[1]{\left\lVert#1\right\rVert}

\usepackage{xspace}
\newcommand{\std}{\textsc{Standard}\xspace}
\newcommand{\stds}{\textsc{Std.}\xspace}
\newcommand{\soft}{\textsc{Soft}\xspace}
\newcommand{\fused}{\textsc{Fused}\xspace}
\newcommand{\context}{C}

\usepackage{listings}
\usepackage{xcolor}
\usepackage{wrapfig}
\lstdefinestyle{lfd}{
  language=Python, basicstyle=\ttfamily\scriptsize,
  commentstyle=\itshape\color{gray}, keywordstyle=\bfseries,
  columns=fullflexible, keepspaces=true, frame=single,
  framerule=0.3pt, xleftmargin=14pt, xrightmargin=4pt,
  numbers=left, numberstyle=\tiny\color{gray}, numbersep=6pt,
  aboveskip=2pt, belowskip=2pt, showstringspaces=false}
\hypersetup{%
    pdfborder = {0 0 0},
    colorlinks,
    citecolor=blue,
    linkcolor=blue,
}

\title{
\centering
Full-bandwidth Transformer
}

\author{
\centering
{\name Xi Wang\textsuperscript{1,$\dagger$,*},
\name Ziyang Cai\textsuperscript{2,$\dagger$},
\name Zheng Zhan\textsuperscript{3},
\name Harry Dong\textsuperscript{3},
\name Ying Fan\textsuperscript{3},\\[0.5em]
\name Gustavo de Rosa\textsuperscript{3},
\name Tim Pearce\textsuperscript{3},
\name John Langford\textsuperscript{3,*}
}
\\[0.5em]
{\normalfont
\textsuperscript{1}Johns Hopkins University
\quad
\textsuperscript{2}Princeton University
\quad
\textsuperscript{3}Microsoft
}
}

\def\month{MM}  
\def\year{YYYY} 
\def\openreview{\url{https://openreview.net/forum?id=XXXX}} 

\begin{document}

\maketitle
\begingroup
\renewcommand{\thefootnote}{}
\footnotetext{
\textsuperscript{*}Correspondence to Xi Wang
<\href{mailto:xwang457@cs.jhu.edu}{xwang457@cs.jhu.edu}>,
John Langford <\href{mailto:jcl@microsoft.com}{jcl@microsoft.com}>. $\dagger$ Work done during an internship at Microsoft AI Frontiers.
Code available at \url{https://github.com/xidulu/Full-bandwidth-transformer}
}
\endgroup

\begin{abstract}
Autoregressive transformers compute along two axes: horizontally across generated tokens, and vertically through model depth. Dense attention gives each token broad horizontal access to the past, but the vertical feedback channel between decoding steps remains narrow: only the sampled token returns to the bottom of the stack, while the top-layer hidden state is discarded. We introduce the \emph{full-bandwidth transformer}, which widens this channel with \emph{latent feedback}: at each decoding step, the previous top-layer hidden state is fused with the sampled token embedding through a gated linear unit and fed back as the next input. Latent feedback lets non-verbalized computation re-enter the stack with a renewed depth budget, while preserving the standard transformer architecture, KV cache, and language-modeling objective. To train full-bandwidth transformers without losing parallel teacher forcing, we use a scheduled multi-pass objective that introduces latent feedback late in pretraining and mixes a small fraction of deeper feedback passes for stability. We train 1B-parameter full-bandwidth transformers up to 400B tokens and find that latent feedback improves validation loss, 5-shot language-model evaluation, math and coding generation, and instruction-tuned performance. With negligible per-token decoding overhead, full-bandwidth transformers match or approach standard transformers trained with roughly $1.5\times$ more tokens, and manage to produce shorter reasoning when no off-policy templates are provided.
\end{abstract}

\section{Introduction}

\begin{figure}[!t]
    \centering
    \includegraphics[width=.9\linewidth]{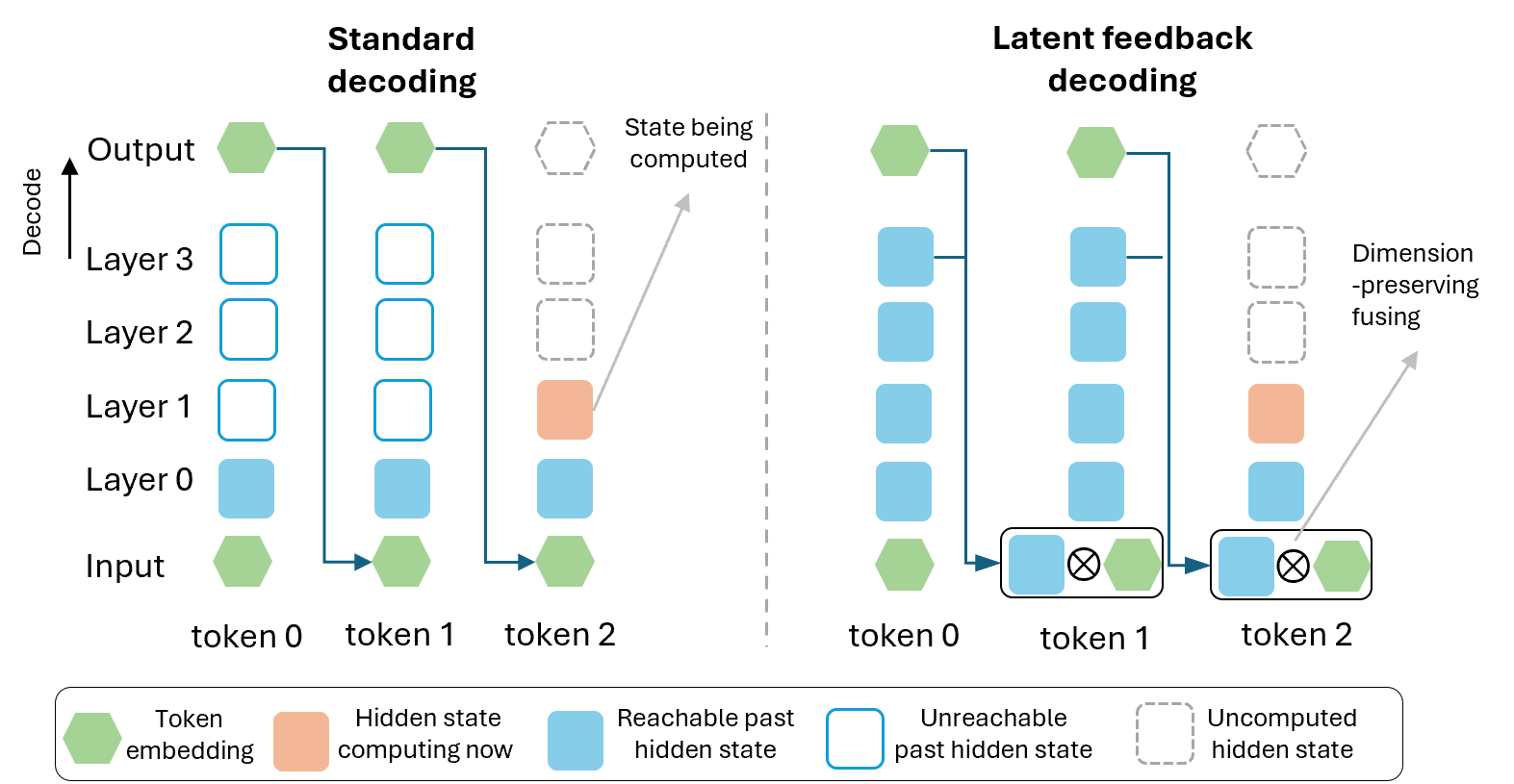}
    \caption{
    \textbf{Standard decoding vs.\ latent feedback decoding.}
    \textbf{Left}: In a standard transformer, the current state can access only lower-layer past states (blue); deeper past states (white) are unreachable, and the only inter-step feedback is the sampled token embedding (green).
    \textbf{Right}: A \emph{full-bandwidth transformer} uses \emph{latent feedback}, fusing the previous top-layer hidden state with the sampled token embedding through a dimension-preserving gate ($\otimes$, Eq.~\eqref{eq:glu_cross}) and feeding it back as the next input. This returns full hidden-state information to the bottom of the stack, making the past as processed by all layers accessible to subsequent computation.
    }
    \label{fig:flow_chart}
\end{figure}

Autoregressive transformers contain a feedback loop: the token sampled at step $t-1$ becomes the input at step $t$ (Fig.~\ref{fig:flow_chart}, left).
This loop is what lets chain-of-thought decoding~\citep{wei2022chain} perform computation whose depth grows with the number of generated tokens~\citep{li2024chain}.
But measured as a communication channel, the loop is narrow:
Decoding compresses the model's entire top-layer state, a $D$-dimensional vector, down to a single symbol carrying at most $\log_2|V|$ bits.
Non-verbalized computation is not erased, intermediate activations persist in the KV cache, but it is \emph{depth-frozen}: a state produced at layer $\ell$ is readable only by layers above $\ell$, so it can never return to the bottom of the stack for further processing, and the deepest, top layer's output is never cached. Verbalization is thus the only channel for information to re-enter the bottom layer and receive fresh computation, at the cost of being squeezed through a single token, forcing the model either to spend tokens narrating its intermediate state or recompute that state from scratch at every position.

In this work, we propose the \emph{full-bandwidth transformer} where we widen this channel to its full width. 
In particular, we introduce latent feedback decoding, which fuses the previous top-layer hidden state with the sampled token's embedding during decoding, and feeds the result back as the next input (Fig.~\ref{fig:flow_chart} right, Sec.~\ref{sec:latent_feedback_decoding}).
We call a transformer capable of decoding this way a full-bandwidth transformer, since its inter-step feedback now carries the entire hidden state rather than a thin token.
The sampled token is retained, so the model still produces ordinary text and can be flexibly trained with standard supervised language modeling losses; what changes is that the feedback is no longer limited to the token's identity.
By design, this affords two things standard decoding lacks: (i) non-verbalized state can re-enter the bottom of the stack with a renewed depth budget and be processed further across steps, rather than staying frozen in the cache at the level where it was produced; (ii) every layer, including the shallowest, sees the past as processed by the \emph{full} stack, not only by the layers beneath it; 
Crucially, these come with almost no architecture changes and extra serving cost: the fusion adds two matrix multiplications per generated token, attention and the KV cache are untouched, and prefill is run either once or, optionally, twice for better performance.

The obstacle is training. A pretrained model has never seen hidden states in its input, so latent feedback cannot simply be switched on at inference; and the recurrence it defines is sequential over positions, so training on it directly would forfeit the parallel teacher forcing that makes transformers efficient to train.
We resolve this with a \emph{multi-pass} regime (Sec.~\ref{sec:recur_training}): each pass shifts the previous pass's hidden states one position rightward, fuses them with the token embeddings, and re-runs the stack in parallel across all positions, so sequentiality is paid across multiple forward passes rather than across the sequence.
To make this practical at large scale, we adopt a \emph{progressive schedule}, spending the bulk of training (e.g. first 75\%) on the ordinary single-pass objective such that the run can start from a standard pretraining checkpoint and introduces extra feedback passes only late; 
Empirically, we find the schedule's composition matters in an unexpected way: training with two feedback passes alone produces a recurrence that \emph{diverges} once rolled past its trained depth, whereas mixing in as little as 3\% three-pass batches turns the learned map into a stable attractor, which stays stable beyond the trained depth (Fig.~\ref{fig:length_extrapolation}).


Empirically, full-bandwidth transformers show performance equivalent to substantially more training data.
Utilizing two or three forward passes for prefill, the model matches standard transformer baselines trained on twice the tokens in both validation loss and multiple-choice accuracy (Fig.~\ref{fig:loopy}).
On free-form generation (Fig.~\ref{fig:base_model_gen_eval})---GSM8K, HumanEval, MBPP---latent feedback improves over standard decoding  under the \emph{same} weights, matches the $2\times$-token baselines, and on some tasks approaches baselines trained with up to $5\times$ the tokens; the gains carry over through long-context extension and instruction tuning (Table~\ref{tab:if_results}).
Additionally, latent feedback often yields shorter reasoning traces, (Fig.~\ref{fig:rollout_length_and_acc} and \ref{fig:qualitative-decimal}), the behavior the widened channel predicts, with compute state stored in the latents instead of being verbalized by tokens.


\section{Background}\label{sec:background}

Given a vocabulary of size $|V|$ and a $D$-dimensional residual stream, a decoder-only LLM, $f_\theta(\cdot)$, maps an input sequence of $T$ tokens, with embeddings $\{\embd_t\}_{t=1}^{T}\in\mathbb{R}^{T\times D}$, through $L$ attention--MLP blocks. The final-layer hidden states $\{\bh_t^{L}\}_{t=1}^{T}$ are then projected and decoded into the next-token
\begin{equation}
\bh_t^{L} = f_\theta(\embd_t;\, \context), \qquad \embd_t \leftarrow \mathrm{Decode}\!\left(\bh_{t-1}^{L}\right), \quad \context = \embd_0, \embd_1, \ldots, \embd_{t-1}.
\label{eq:standard-decode}
\end{equation}


\paragraph{Bandwidths of a model's horizontal axis vs. vertical axis.} It is useful to separate the horizontal axis (across positions) from the vertical axis (across depth). \emph{Horizontally}, dense attention is effectively full-bandwidth: when generating token $t$, the layer-$\ell$ state $\bh_t^{\ell}$ can read the cached representations of every earlier position.
\emph{Vertically}, access is restricted: $\bh_t^{\ell}$ cannot read any deeper past state $\bh_{t'}^{\ell'}$ with $t'<t$ and $\ell'\ge\ell$ (Fig.~\ref{fig:flow_chart}, left).
Formally, the states reachable when computing position $t$ at layer $\ell$ are
\begin{equation}
\mathcal{R}_{\text{std}}(t,\ell) = \big\{(t',\ell') : t'<t,\; \ell'<\ell\big\},
\qquad
\bigl\lvert \mathcal{R}_{\mathrm{std}} \bigr\rvert = \Theta(t\ell),
\label{eq:std-reach}
\end{equation}
so a shallow layer of a new token sees only a \emph{partially processed} view of the past, even though the deeper, more fully processed states of those same positions have already been computed and sit in the KV cache.
Past computation produced at layer $\ell$ is readable only to layers above $\ell$ and can never be routed back down for further processing.
This is the narrow vertical channel Sec.~\ref{sec:latent_feedback_decoding} widens.

Importantly, this depth-wise dependency constraint is also what lets transformers train in parallel across positions: sequential computation is required only across layers, not across tokens. At decoding time, however, generation is already sequential over tokens, so the constraint buys nothing, opening the door to richer dependencies on past hidden states, which we develop next.

\section{Widening the bandwidth with latent feedback decoding}
\subsection{Latent feedback decoding}\label{sec:latent_feedback_decoding}
The central technique in full-bandwidth transformers is latent feedback decoding, which feeds the previous top-layer hidden state back into the input. At step $t$,
\begin{equation}
    \bh_{t}^L = f_\theta\!\left(\cross{\embd_t}{\bh_{t-1}^L};\; \context\right), ~
    \textrm{where}~\embd_t \leftarrow \mathrm{Decode}\!\left(\bh_{t-1}^L\right),\; \context=\embd_1, \cross{\embd_2}{\bh_0^1},\ldots, \cross{\embd_{t}}{\bh_{t-1}^L}
    \label{eq:latent-feedback}
\end{equation}
where $f_\theta$ is the $L$-layer transformer stack, $\cross{\cdot}{\cdot}$ fuses the sampled token's embedding with the previous latent state, and $\context$ is the past context (the KV cache of all earlier positions). 

The fusion $\otimes$ is a gated linear unit:
\begin{equation}\label{eq:glu_cross}
\cross{\embd_t}{\bh_{t-1}} = \bWU \bh_{t-1} \odot \sigma(\bWG \embd_{t}),
\end{equation}
with $\bWU, \bWG \in \R^{D\times D}$. The asymmetry is deliberate: the hidden state occupies the value pathway, while the token embedding enters only as a multiplicative gate. 
We hypothesize that a symmetric fusion such as $\bWG\embd_t + \bWU\bh_{t-1}$ would leave a shortcut open where the model could ignore the $\bh$ and recover the plain token input. 
That shortcut is tempting when training starts from a standard checkpoint where ignoring $\bh$ could give lower loss.
However, in small scale ablation study, we do not observe evidence of such shortcut under symmetric fusions (App.~\ref{app:symmetric_gate}).

\paragraph{Latent feedback is free to serve.}
The added inference cost is independent of context length and model depth.
The state $\bh_{t-1}^L$ is already computed during standard decoding, so the only extra work is two $D\times D$ matrix multiplications, negligible against a forward pass through $L$ blocks ($< 1\%$). Since fusion preserves the input dimension $D$, the architecture, KV-cache layout, and serving stack are untouched, and the decoding loop changes by two lines (Fig.~\ref{fig:pseudocode}, right). The scheme is also vLLM-compatible: we store top-layer states in a dedicated buffer, adapting the mechanism used by multi-token-prediction implementations (Appendix~\ref{sec:vllm}). The end-to-end latency results under vLLM are shown in Appendix~\ref{sec:latency}.

\paragraph{Latent feedback improves computational accessibility.}
The main benefit of latent feedback decoding comes from lifting the depth restriction of Eq.~\eqref{eq:std-reach}.
Now every layer, including the lowest, can access top-layer information,
\begin{equation}
\label{eq:reach-lf}
\mathcal{R}_{\mathrm{lf}}(t,\ell)
\;=\; \bigl\{\, (t',\ell') \;:\; t' < t,\; 0 \le \ell' \le L \,\bigr\},
\qquad
\bigl\lvert \mathcal{R}_{\mathrm{lf}} \bigr\rvert = \Theta(TL),
\end{equation}
shown in Fig.~\ref{fig:flow_chart} (right).
The improved accessibility is empirically verified in synthetic settings in Sec.~\ref{sec:state_tracking} and leads to improved performance in free-form generation tasks (Sec.~\ref{sec:gen_eval}).

\paragraph{Latent feedback adds draft space.}
Latent feedback also provides a continuous scratchpad, relieving the pressure to verbalize intermediate state. State maintenance moves from the sequence axis alone to the depth axis as well: intermediate results can be updated recurrently over time, through $\bh$ along the stack rather than only by extending the token sequence.
This predicts the possibility of shorter rollouts, for which Sec.~\ref{sec:concise_reasoning} provides preliminary evidence.

\paragraph{What latent feedback does NOT provide.} We provide two important clarifications:
\begin{itemize}[leftmargin=*, topsep=0pt, parsep=0pt, itemsep=1.5pt]
    \item \textbf{No mutable register.} RNNs and state-space models overwrite a compressed state at each step. Latent feedback is recurrent in form, but past states persist in the KV cache rather than being overwritten, so every earlier state remains readable by the current token.
    \item \textbf{No added asymptotic depth at decoding time.} Latent feedback does not change the asymptotic serial depth of decoding: with or without it, each step has a depth-$\mathcal{O}(L)$ graph, so $T$ tokens cost $\mathcal{O}(TL)$. What changes is the \emph{bandwidth} of the path, with a verbal channel and a continuous channel now \emph{evolving in parallel}.
    Note that a full-bandwidth transformer can further increase the depth at prefilling time through a  multipass prefill, which we will introduce in the following section.
\end{itemize}
 
\subsection{Parallel multi-pass training for latent feedback decoding}\label{sec:recur_training}

\begin{figure}[t]
\begin{minipage}[t]{0.48\textwidth}
\begin{lstlisting}[style=lfd, caption={Training: one step with $k$ passes.}]
def glu_cross(h, e): # [T,D],[T,D]->[T,D]
    return (h @ W_u) * sigmoid(e @ W_g)
 
e = embed(tokens)  # [T, D]
h = model(e)       # pass 1 (standard)
loss = ntp_loss(h)
for _ in range(k - 1): # parallel in T
    x = glu_cross(shift_right(h), e)
    # random plain prefix
    x = prefix_mixin(x, e)
    h = model(x)
    loss += ntp_loss(h) / (k-1)
\end{lstlisting}
\end{minipage}\hfill
\begin{minipage}[t]{0.52\textwidth}
\begin{lstlisting}[style=lfd, caption={Inference (\soft); uncommenting line 2-3 gives \fused; line 7 shows the \std-decoding input).}]
h = model(embed(prompt)) # prefill, h: [T, D]
#h = model(
#   glu_cross(shift_right(h), embed(prompt)))
tok = sample(lm_head(h[-1]))
h_prev = h[-1]
while not done:             # decode
    x = glu_cross(h_prev, embed(tok))
    # standard decoding: x = embed(tok)
    h_prev = model_step(x, kv_cache)
    tok = sample(lm_head(h_prev))
\end{lstlisting}
\end{minipage}
\caption{\textbf{Latent feedback in pseudo-code (simplified).} Training (left) pays sequentiality across $k$ passes, each parallel over positions. Inference (right) differs from standard decoding by a single line (line 6 vs.\ the commented line 7): the input is the fused state rather than the token embedding alone, reusing the state previously used for decoding. Fig.~\ref{fig:pseudocode_complete} in the appendix shows the full training and inference procedure with additional stability techniques such as input normalization included.
}
\label{fig:pseudocode}
\end{figure}

\begin{figure}[h]
    \centering
    \includegraphics[width=0.9\linewidth]{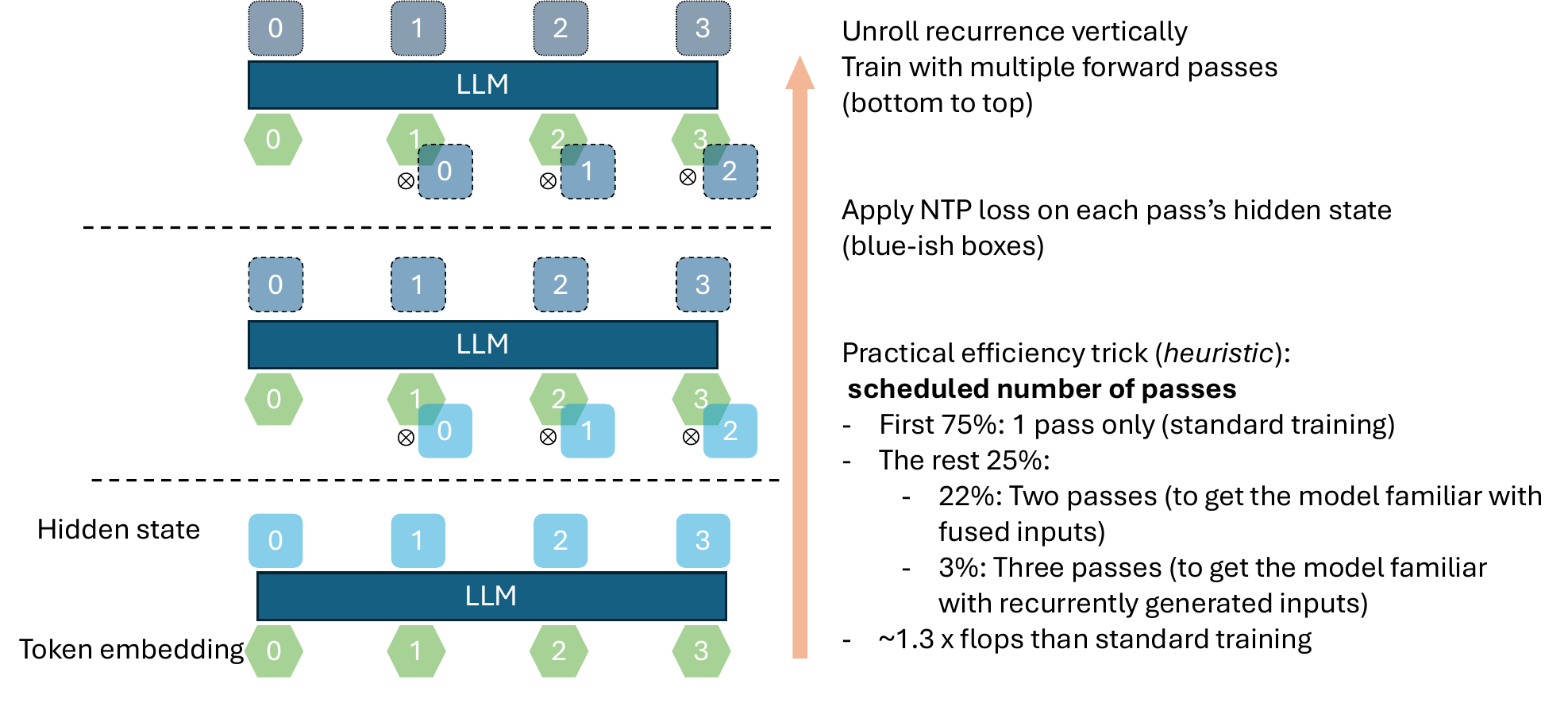}
    \caption{Illustration of multi-feedback pass training}
    \label{fig:train_diagram}
\end{figure}


At decoding time, latent feedback unrolls over generated positions. Let $\bu_t$ be the input actually fed to the transformer stack at position $t$. The input sequence looks like
\begin{equation}
    \bu_1 = \embd_1,\quad
    \bh_1 = f_\theta(\bu_1; C_1);
    \qquad
    \bu_t = \cross{\embd_t}{\bh_{t-1}},\quad
    \bh_t = f_\theta(\bu_t; C_t),
    \quad t \ge 2.
\label{eq:decode-unroll}
\end{equation}
Here $\cross{\cdot}{\cdot}$ is the gated fusion of
Eq.~\eqref{eq:glu_cross}, and $C_t$ is the KV cache over the previous inputs
$\bu_{1:t-1}$. Thus the stack sees the input sequence
$
    \embd_1,\ 
    \cross{\embd_2}{\bh_1},\ 
    \cross{\embd_3}{\bh_2},\ 
    \cross{\embd_4}{\bh_3},\ldots
$
rather than plain embeddings alone. Since a standard next-token-prediction
model is trained only on plain token embeddings in this slot, full-bandwidth
transformers must be trained on these latent-feedback inputs as well.

The exact recurrence of Eq.~\eqref{eq:decode-unroll} is sequential in $t$,
so training on it directly requires RNN-style backpropagation through time (BPTT), which breaks the parallel teacher forcing. 
We instead adopt a multi-forward-pass approximation. For each position in the sequence, we compute the top-layer state several times, writing $\bh_t\pass{k}$ for the state at position $t$ on pass $k$ (the layer superscript $L$ is omitted throughout this section):
\begin{align}
\bh_t\pass{1} &= f_\theta(\embd_t;\, \context\pass{1}),
  & \context\pass{1} &= \embd_1, \ldots, \embd_{t-1}, \label{eq:pass1}\\
\bh_t\pass{2} &= f_\theta\!\big(\cross{\embd_t}{\bh_{t-1}\pass{1}};\, \context\pass{2}\big),
  & \context\pass{2} &= \embd_1,\, \cross{\embd_2}{\bh_1\pass{1}},\, \ldots,\, \cross{\embd_{t-1}}{\bh_{t-2}\pass{1}}, \label{eq:pass2}\\
  &\qquad\ldots \nonumber \\
\bh_t\pass{k} &= f_\theta\!\big(\cross{\embd_t}{\bh_{t-1}\pass{k-1}};\, \context\pass{k}\big),
  & \context\pass{k} &= \embd_1,\, \cross{\embd_2}{\bh_1\pass{k-1}},\, \ldots,\, \cross{\embd_{t-1}}{\bh_{t-2}\pass{k-1}}. \label{eq:pass3}
\end{align}
The first pass is the ordinary no-feedback forward pass ($\bh_t\pass{1} \equiv \bh_t$); each subsequent pass shifts the previous pass's states one position rightward, fuses them with the token embeddings, and re-runs the full stack in parallel across all positions, since every state it requires was completed in the previous pass.
We then apply the standard teacher-forced next-token-prediction loss
to the outputs of \emph{every pass}.
Retaining the first-pass loss preserves the model's no-feedback mode of operation, which is what processes the prompt at inference time.
We \emph{do not detach the gradient}, so the loss from later passes backpropagates into earlier passes' latent states, acting as an auxiliary objective; this does increase the memory footprint. The overall objective is
\begin{equation}
\mathcal{L}^{K}(\theta)
= \underbrace{\sum_{t=1}^{T} -\log p_\theta\!\left(x_{t+1} \mid \embd_{1:t}\right)}_{\text{standard NTP objective}}
\;+\; \lambda \,\frac{1}{K-1}\sum_{k=2}^{K} \sum_{t=1}^{T}
-\log p_\theta\!\left(x_{t+1} \mid \bu_{1:t}\pass{k}\right),
\label{eq:recur-loss}\vspace{-1mm}
\end{equation}
where $\bu_{1:t}\pass{k} = \embd_1,\, \cross{\embd_2}{\bh_1\pass{k-1}},\, \ldots,\, \cross{\embd_t}{\bh_{t-1}\pass{k-1}}$ are the pass-$k$ fused inputs of Eqs.~\eqref{eq:pass2}--\eqref{eq:pass3}. In all experiments we set $\lambda = 1$ without any tuning.

The pseudocode and diagram is shown in Fig.~\ref{fig:pseudocode} left and Fig.~\ref{fig:train_diagram}.
We refer to this scheme as \emph{temporal parallelism}, following a common strategy for parallelizing recurrent computation during training~\citep{zeng2025ponderlm, cai2026t, huang2026latent, liu2026maglev, danieli2025pararnn}.
After $k$ passes, a top-layer state from position $t$ can affect the input at positions up to $t+k-1$, so $k$ passes train the feedback transition over a horizon of $k-1$ token steps. Training thus pays sequentiality across passes rather than across positions.


\paragraph{Feedback-pass scheduling.}
At decoding time, the feedback loop unrolls indefinitely, so the trained map must remain stable under many more self-compositions than any training budget can simulate; yet running many passes throughout training is prohibitively expensive, since each pass multiplies the cost of the run. Scheduling the number of forward passes---how many, and when---is therefore central to making latent-feedback training practical.

\begin{wrapfigure}[18]{r}{0.43\textwidth}
    \centering
    \vspace{-1.5em}
    \includegraphics[width=\linewidth]{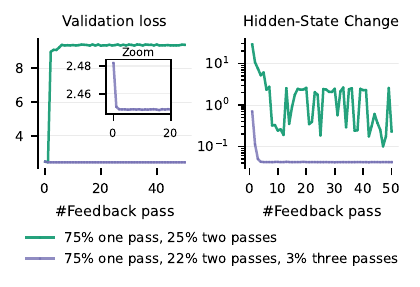}\vspace*{-4mm}
    \caption{
    \textbf{Three-pass training stabilizes long-horizon latent feedback.}
    With only single- and two-pass batches, validation loss and hidden-state updates become unstable beyond the training horizon. Adding a small fraction of three-pass batches yields stable loss and bounded hidden-state updates ($\textstyle \E{t}{\|\bh_t\pass{k}-\bh_t\pass{k-1}\|}$).
    }
    \label{fig:length_extrapolation}
    \vspace{-0.7em}
\end{wrapfigure}

\emph{How many passes.}
We choose the number of passes based on \emph{empirically } checking whether the iterated feedback map reaches a stable state.
Our experiments show that full-bandwidth transformer takes only 3 forward passes to reach a stable state, where further forward passes do not change the hidden state or validation loss.
We hypothesize there are two factors contributing to the fast convergence:
(1). Each pass's input is only partially changed: The token embedding is fixed, and only the hidden-state is changing; 
(2). We apply NTP supervision on the output of each pass (analogous to intermediate states in loop transformer), a technique critical for stability and faster convergence of loop transformer suggested by recent works \citep{zhu2025scaling, sharma2026dense, shomali2026loopmtp}.


\emph{When to introduce feedback passes.}
Feedback passes are expensive, so most of training uses the standard single-pass objective, and we add extra passes progressively in the middle of training, beginning with two-pass batches, and later with a small fraction of batches with more passes.
This lets the run begin from an ordinary mid-trained checkpoint, spend the bulk of its compute on standard teacher forcing, and pay the extra feedback-pass cost only in the end.

Fig.~\ref{fig:length_extrapolation} illustrates the feasibility of the scheduling.
We studied a 1B model trained on 200B tokens. A model trained with only single- and two-pass batches (75\% single-pass, 25\% two-pass; green) performs well at the trained depth but fails to extrapolate.
Adding 3\% three-pass batches (75\% single-pass, 22\% two-pass, 3\% three-pass; blue) changes the behavior: validation loss remains flat through $30$ feedback steps, and the hidden-state change decays to a small plateau. 
The same extrapolation behavior carries over to inference: hundred-token rollouts show no sign of breakdown, and we observe similar stability under $k=1000$ feedback passes (Fig.~\ref{fig:length_extrapolation_100}).
However, the design of scheduling is mainly based on empirical experiments and failure of convergence does occur when pursuing extreme efficiency (App.~\ref{sec:schedule_ablation}). The exact reason for converging in 3 passes is also unclear and has also been empirically observed by other works~\citep{huang2026latent, danieli2025pararnn}.

Lastly, during actual training, we included additional techniques for stability.
\paragraph{Prefix mixin.} A distribution mismatch remains between multi-pass training and inference. At decoding time a sequence is heterogeneous: prompt positions carry plain token embeddings (processed by a single prefill pass), while generated positions carry fused inputs. In the passes of Eqs.~\eqref{eq:pass2}--\eqref{eq:pass3}, by contrast, \emph{every} position beyond the first is fused. A model trained only on fully-fused passes therefore encounters an out-of-distribution boundary at inference, precisely where the prompt ends and generation begins. To close this gap we apply a \emph{prefix mixin}: in each pass beyond the first, we sample a random prefix length $p$ and revert positions $t \leq p$ to plain embeddings, fusing only the suffix.
Training thus covers sequences that switch from plain to fused inputs at an arbitrary point, i.e. the structure of single-prefill inference. 
Alternatively, the prompt itself can be run through a second, fused prefill pass so that all positions match the fused distribution; the mixin removes the need for this, but we support both, corresponding to the ``identical or doubled prefill'' overhead stated in the abstract.


\paragraph{Stability recipes for long feedback horizons.}
At inference time, latent feedback may be applied for hundreds or thousands of generated tokens, far beyond the few feedback passes used during training. We therefore use several lightweight stabilization techniques to keep the feedback map well behaved under long self-composition.
\begin{itemize}[leftmargin=*, topsep=0pt, parsep=0pt, itemsep=1.5pt]
    \item \textbf{Stationary hidden-state scale.}
    We keep the magnitude of carried state $\bh_t^L$ bounded as feedback is repeatedly applied. To prevent the top-layer state norm from growing with depth, we use depth scaling~\citep{yang2024tensor,noci2022signal} so that $\norm{\bh_t^L} \sim \mathcal{O}(1)$ rather than $\mathcal{O}(L)$, as can occur in a standard pre-norm model.
    We also apply RMSNorm to the fused input $\cross{\embd_t}{\bh_{t-1}^L}$ before feeding it into the model.
    \item \textbf{Shared input basis with weight tying.}
    The model processes two types of inputs: plain token embeddings during standard prefill, and fused hidden-state/token inputs during latent-feedback decoding.
    We therefore encourage the embedding space and top-layer hidden-state space to remain in a compatible basis by tying the weights of the embedding layer and readout layer, reducing the burden on the fusion weights to learn a large corrective rotation between the two input distributions.
    \item \textbf{Noise regularization.}
    During training, we add small jitter noise to the carried hidden state before fusion,
    \begin{equation}\label{eq:jitter_noise}
        \bh_t^L
        =
        f_\theta\!\left(
            \cross{\embd_t}{\bh_{t-1}^L + \epsilon};
            \context
        \right),
        \qquad
        \epsilon \sim \mathrm{Uniform}[-\sigma,\sigma]^D .
    \end{equation}
    This exposes the feedback map to a local neighborhood around each training state, making it less sensitive to small deviations that can accumulate over long feedback horizons.
\end{itemize}
The complete pseudo code for training where the tricks are adopted is presented in Fig.~\ref{fig:pseudocode_complete} in the appendix. It is also worth noting that we only ablated the injected noise (Fig.~\ref{fig:jitter_noise_ablate}) but we did not ablate the depth scaling and weight tying. We adopted these techniques in that we hypothesize that they are, motivationally, the right thing to do, which may not be actually necessary empirically.





\section{Experiments}\label{sec:experiments}
To evaluate full-bandwidth transformers, we pretrain 1B-parameter models (architecture described in Appendix~\ref{sec:model_arch}) using the training recipe from Sec.~\ref{sec:recur_training}.
We use NorMuon~\citep{li2026normuon} for matrix parameters with learning rate $1\times 10^{-2}$ and weight decay $0.01$, and Adam for all other parameters with learning rate $5\times 10^{-4}$
and no weight decay.
All runs use a WSD learning-rate schedule \citep{hagele2024scaling, hu2024minicpm} with 200 warmup steps and a 25\% cooldown phase decaying to zero. During cooldown, we add a z-loss \citep{chowdhery2023palm} with coefficient $1\times 10^{-5}$ and decay weight decay together with the learning rate following AdamC~\citep{defazio2025gradients}.
For all experiments, we use a jitter noise with $\sigma=0.02$ (Eq.~\eqref{eq:jitter_noise}) during training.
Models are trained on the data mixture of Phi-4~\citep{abdin2024phi}, with context length 8192. For all runs, we use a global batch size of 300K tokens; except for the 1T-token baseline run, which uses a larger global batch size of 1.2M tokens.
We summarize below in Table \ref{tab:budget_summarization} the exact scheduling and full-bandwidth transformer runs' \emph{token-equivalent compute},
defined as training tokens multiplied by the average number of forward passes per batch. 
\begin{center}
\small
\begin{table}[!h]
\begin{tabular}{lccc}
\toprule
Run & Feedback-pass schedule & Tokens & Token-equivalent compute \\
\midrule
10B  & 100\% three-pass & 10B  & 30B \\
100B & 75\% one-pass, 25\% three-pass & 100B & 150B \\
200B & 75\% one-pass, 22\% two-pass, 3\% three-pass & 200B & 256B \\
400B ($1.28\times$) & 75\% one-pass, 22\% two-pass, 3\% three-pass & 400B & 512B \\
400B ($1.5\times$) & 75\% one-pass, 25\% three-pass & 400B & 600B \\
\bottomrule
\end{tabular}
\caption{Summarization of each experiment run's schedule.}\label{tab:budget_summarization}
\end{table}
\end{center}
Example validation loss dynamics under 400B tokens run, of different passes at different schedules, are shown in Fig.~\ref{fig:training_dynamics} in the appendix. The actual per-step training time under different number of passes and the inference latency are shown in App.~\ref{sec:train_runtime} and App.~\ref{sec:latency}.

\subsection{Fused prefilling improves non-generative performance}
\label{sec:loopy_results}

\begin{figure}[!t]
    \centering
    \includegraphics[width=\linewidth]{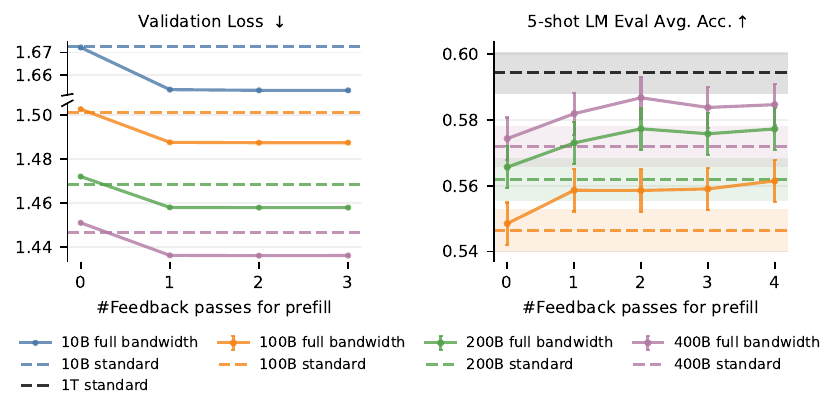}\vspace*{-4mm}
    \caption{
    \textbf{Feedback passes during prefilling improve non-generative performance.} Under a 1B model trained under different number of tokens, re-running prefill with additional fused passes (Eqs.~\eqref{eq:pass2}--\eqref{eq:pass3}) improves both validation perplexity (left) and 5-shot LM Eval accuracy across 10 tasks (right) across training scales;
    Error bars denote one standard error of the mean, obtained by propagating individual per-task standard errors (summing in quadrature and dividing by the number of tasks). The 400B token model refers to the $1.28\times$ extra compute version, the $1.5\times$ compute version shows stronger results, shown in Fig.~\ref{fig:lm_eval_400b_extra} in appendix.
    }
    \label{fig:loopy}
\end{figure}




Latent-feedback training enables a simple form of prefill-time test-time scaling. At evaluation, we apply $k$ additional fused passes over the prompt using Eqs.~\eqref{eq:pass2}--\eqref{eq:pass3}.
These passes augment embeddings with richer context information from top-layer hidden states, 
improving perplexity and downstream accuracy.

Fig.~\ref{fig:loopy} plots validation loss and average 5-shot LM Eval accuracy across RTE, TruthfulQA-MC2, ARC-Easy, ARC-Challenge, BoolQ, PIQA, WinoGrande, OpenBookQA, COPA, and MMLU, as a function of the number of feedback passes applied during prefill.
Step~0 is ordinary prefill using embeddings as input.
Step~$k$ corresponds to $k$ extra forward passes.
Broadly, most of the improvement appears after the top-layer hidden states are first made available at the input. Further passes continue to help, but with diminishing returns.
In addition, \emph{latent feedback training costs little when unused}. At step~0, where the model is evaluated as an ordinary transformer, full-bandwidth transformers give up only a small amount of validation loss relative to the standard baseline, while already improving average LM Eval accuracy. 
Notably, a small amount of prefill-time compute matches substantially larger standard baselines: With two feedback passes, the 100B-token full-bandwidth transformer reaches the 200B-token standard baseline, and the 200B-token full-bandwidth transformer reaches the 400B-token standard baseline.
Lastly, the 400B full bandwidth model refers to the $1.28\times$ version, the $1.5\times$ compute version (Fig.~\ref{fig:lm_eval_400b_extra} in appendix) gives stronger results under more training time budget.

We additionally compare our model with other models of similar parameter scale on 0-shot LM Eval performance, shown in Table~\ref{tab:obs-comparison} in Appendix~\ref{sec:lm_eval_comp}, where we found our model performs on-par or better than models trained under similar or more tokens.
These results imply that full-bandwidth transformers improve on a strong baseline.



\subsection{Latent feedback decoding improves decoding performance}\label{sec:gen_eval}
\begin{figure}[!t]
    \centering
    \includegraphics[width=\linewidth]{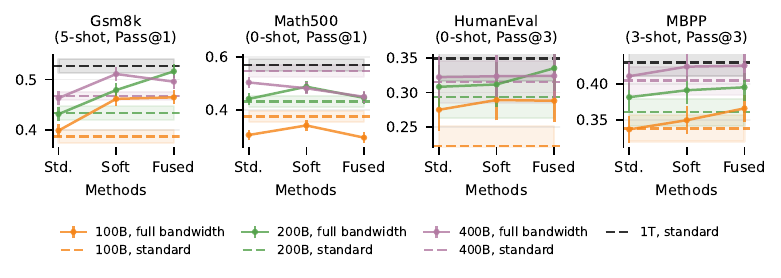}\vspace*{-4mm}
    \caption{
    \textbf{Latent-feedback decoding improves pre-trained base 1B parameter models at fixed decoding cost.}
    We compare the three decoding regimes from Sec.~\ref{sec:gen_eval}.
    Coding results report Pass@3 from 10 rollouts, with temperature selected from $\{0.3,0.5,0.7\}$ separately for each method (for \emph{both} baseline and our model).
    \soft generally improves over \std at identical inference cost, while \fused can yield further gains with \(2\times\) prefill compute and unchanged decoding cost.
    The MATH-500 degradation observed before instruction tuning is reversed after instruction tuning, with the best full-bandwidth regime matching or exceeding the standard baseline at each scale (Table~\ref{tab:if_results}).
    }
    \vspace{-3mm}
    \label{fig:base_model_gen_eval}
\end{figure}

We now evaluate open-ended generation tasks. We compare three decoding regimes:
\begin{itemize}[leftmargin=*, topsep=0pt, parsep=0pt, itemsep=1.5pt]
    \item \std: single-pass prefill; generation uses token embeddings only. This evaluates the full-bandwidth model as an ordinary transformer, and measures the cost of latent-feedback training when the feedback channel is not used at inference.
    \item \soft: single-pass prefill; generation uses latent feedback as in Eq.~\eqref{eq:latent-feedback}. 
    Same per-token overhead as \std except for the two $D\times D$ matrix multiplications.
    \item \fused: the prompt is first processed by an additional fused prefill pass for refinement, as in Eq.~\eqref{eq:pass2}; generation then proceeds as in \soft. 
\end{itemize}
Thus \std and \soft have identical prefill cost, while \fused doubles prefill cost while keeping the same per-token decoding cost as \soft and \std.

\paragraph{Evaluation setting.}
We evaluate on GSM8K~\citep{cobbe2021training}, MATH-500~\citep{lightman2023lets}, HumanEval~\citep{chen2021codex}, and MBPP~\citep{austin2021program}. We report Pass@1 for math and Pass@3 for coding. For coding, Pass@3 is estimated from 10 rollouts per problem, with temperature grid-searched over $\{0.3,0.5,0.7\}$ separately for each decoding regime. We do not use top-$k$ or top-$p$.
The median generation length for each task is shown in Table~\ref{tab:iycj_generation_length} in appendix.
 
\paragraph{Latent feedback decoding improves the base model.}
Fig.~\ref{fig:base_model_gen_eval} compares the three decoding regimes for full-bandwidth models (solid lines) against standard baselines (dashed lines).
Similar to Fig.~\ref{fig:loopy}, the 400B full bandwidth model refers to the $1.28\times$ version and the $1.5\times$ shows consistently better performance (red lines, Fig.~\ref{fig:free_form_400b_extra}).
Across GSM8K and coding tasks, \soft generally improves over \std at the same decoding cost, while \fused can provide further gains with $2\times$ prefill compute.
MATH-500 is the main exception: before instruction tuning, latent feedback does not consistently improve over \std, particularly at larger scale.
On the remaining tasks, full bandwidth models often approach or exceed standard baselines trained with $2$--$5\times$ more tokens, including the 1T baseline on GSM8K and HumanEval.
Finally, Pass@3 improvement on coding tasks suggests that conditioning generation on recurrent hidden states does not reduce sampling diversity.
 
\paragraph{The improvement carries over through instruction tuning.}
We further apply long-context extension (12B tokens) from 8K to 32K and instruction tuning (6B tokens) for models of different pre-train budgets and then evaluate without few-shot examples.
Because these stages are much shorter than pretraining, we train them with \emph{three} forward passes throughout rather than using the pretraining feedback-pass schedule\footnote{Baseline standard models are trained on the same set and number of tokens since there is no extra corpus available for spending more compute on.}. 
Results are shown in Table~\ref{tab:if_results}, where the gains from latent feedback largely persist after instruction tuning.
\soft and \fused are generally competitive with or improve upon STANDARD, and the best latent-feedback regime outperforms the matched-token standard baseline on all four tasks at every reported scale.
In particular, the MATH-500 degradation observed in the base models disappears after instruction tuning.
At 400B (1.5$\times$), SOFT also exceeds the 1T-token standard baseline on MATH-500 ($49.0$ vs.~$47.4$).
Table~\ref{tab:if_results_extended} reports additional instruction-following and writing benchmarks, where we observe little difference between approaches.
Finally, increasing the number of feedback passes for \fused does not yield further gains (Fig.~\ref{fig:num_pass_ablation}).


\begin{table}[t!]
\centering
\small
\setlength{\tabcolsep}{3.2pt}
\begin{tabular}{lcccccccccccc}
\toprule
& \multicolumn{3}{c}{FBT, 200B}
& \multicolumn{3}{c}{FBT, 400B ($1.28\times$)}
& \multicolumn{3}{c}{FBT, 400B ($1.5 \times$)}
& \multicolumn{3}{c}{Standard transformer} \\
\cmidrule(lr){2-4} \cmidrule(lr){5-7} \cmidrule(lr){8-10} \cmidrule(lr){11-13}
Task
& \stds & \soft & \fused
& \stds & \soft & \fused
& \stds & \soft & \fused
& 200B & 400B & 1T \\
\midrule
GSM8K
& 64.52 & \textbf{67.93} & 67.55
& 67.90 & 71.00 & \textbf{71.80}
& 68.08 & 70.74 & \textbf{73.09}
& 62.93 & 68.39 & 70.13 \\

MATH-500
& 43.80 & \textbf{45.60} & \textbf{45.60}
& 46.00 & 45.40 & \textbf{48.40}
& 45.40 & \textbf{49.00} & 48.80
& 42.40 & 46.40 & 47.40 \\

HumanEval
& 42.54 & 45.06 & \textbf{45.92}
& 46.50 & 47.20 & \textbf{47.60}
&45.69 & 46.35 & \textbf{48.77}
& 37.16 & 44.85 & 50.01 \\

MBPP
& 38.39 & 39.80 & \textbf{41.22}
& 40.50 & 40.60 & \textbf{41.70}
& 40.23 & 41.36 & \textbf{42.34}
& 38.61 & 40.28 & 41.93 \\
\bottomrule
\end{tabular}
\vspace{-2mm}
\caption{
\textbf{Latent-feedback gains carry over through instruction tuning.}
We evaluate full-bandwidth transformers after long-context extension and instruction tuning, using no few-shot examples.
Scores are percentages.
For math tasks, we report Pass@1; for coding tasks, we report Pass@3 estimated from 10 rollouts per problem, selecting the best temperature from $\{0.3,0.5,0.7\}$ for each setting.
Bold indicates the best decoding regime within each full-bandwidth training scale.
Results on instruction following and creative writing benchmarks are presented in table~\ref{tab:if_results_extended} in the appendix, where we do not find the performance varying significantly with decoding regime.
}
\label{tab:if_results}
\end{table}


\subsection{Latent feedback enables more concise reasoning}\label{sec:concise_reasoning}

\begin{wrapfigure}[9]{r}{0.48\textwidth}
    \centering
    \vspace{-1.5em}
    \includegraphics[width=\linewidth]{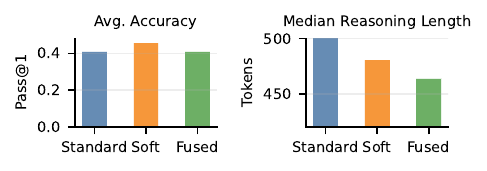}
    \caption{
        Reasoning length and accuracy on Math500 from the 200B run
        (green line, Fig.~\ref{fig:base_model_gen_eval}).
    }
    \label{fig:rollout_length_and_acc}
\end{wrapfigure}

On the base model and 0-shot setting, \soft\ decoding often produces shorter reasoning traces than \std\; 
Fig.~\ref{fig:rollout_length_and_acc} shows the median rollout length on a 200B token model (Fig.~\ref{fig:rollout_length_and_acc_extended} in appendix observe same patterns observed in other runs).
Fig.~\ref{fig:qualitative-decimal} in Appendix~\ref{sec:extra_model_outputs} shows example shortened rollouts.
This is consistent with the behavior predicted by the widened feedback channel: Intermediate computation that would otherwise need to be represented through generated tokens may instead be carried in the recurrent hidden state, reducing the number of tokens needed to reach an answer.

Notably, the effect is less pronounced after instruction tuning or given few shot examples. We hypothesize the cause as external reasoning examples being off-policy with respect to latent-feedback decoding: the target traces were produced by standard token-by-token reasoning, so fitting them re-imposes the fully verbalized style regardless of what the state can carry. On-policy post-training~\citep{ma2026mopd} under latent feedback may preserve the conciseness, which we leave to future work. 

\subsection{Synthetic data experiments and ablation study} 
\paragraph{State extraction and tracking on synthetic data} In appendix \ref{sec:state_tracking}, we generate synthetic sequences and probe the ending token's residual about states determined by tokens at the beginning of the sentence. When prefilled with latent feedback at the last few tokens, shallow layers show significantly higher accuracy than under standard prefilling in shallow layers, since latent feedback exposes deep layer states, which include global information, to shallow ones.

\paragraph{Ablation study} In appendix \ref{sec:ablation_study}, we inspect the learned gating weights, $\bWU, \bWG$ from Eq.~\eqref{eq:glu_cross}, where we observe both pathways contributing non-trivially.
We confirm the empirical performance benefit of \emph{not} detaching hidden state at each pass and the role of injected jitter noise (Eq.~\eqref{eq:jitter_noise}). We also tested alternate scheduling where multi-forward pass steps interleave with one-pass steps, which prevents continuing from a mid-train checkpoint and shows suboptimal performance compared with having each stage in bulk. 

\section{Efficient Replay of Latent-Feedback Trajectories}

\subsection{Multi-feedback pass for KV cache replay}
In actual deployment, a KV cache often needs to be reconstructed from a stored token sequence, for example when resuming an old request.
For tokens generated from \soft, ordinary prefill with embeddings as input does not reproduce the original cache: each generated token was processed using both its token embedding and the preceding token's recurrent hidden state.
These hidden states are not directly available from the token sequence. Replaying the \soft recurrence sequentially can recover them, but requires one dependent forward step per generated token.

The KV cache can be approximately reconstructed using multiple parallel feedback passes, similar to Eq.~\eqref{eq:pass1} to Eq.~\eqref{eq:pass3}, and we keep the final pass's cache as the final KV cache. Importantly, such an approximation is parallelizable over all tokens.

\begin{figure}[!h]
    \centering
    \includegraphics[width=\linewidth]{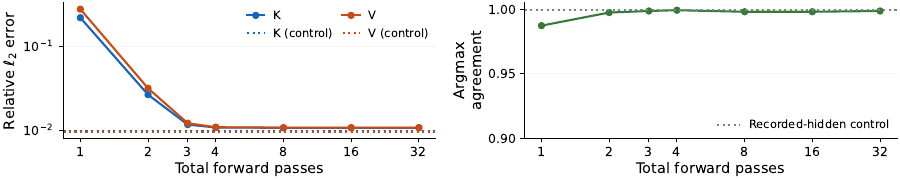}
    \caption{
\textbf{Three to four forward passes recover the KV cache accurately.} 
\textbf{Left:} Relative $\ell_2$ error of reconstructed key (K) and value (V) caches as a function of the total number of forward passes, measured against the caches produced during sequential \soft decoding.
\textbf{Right:} Fraction of positions for which the reconstructed cache yields the same next-token argmax as the sequential-decoding cache.
Dotted lines show the reference error obtained by directly using the actual recurrent hidden states recorded during decoding. 
}
    \label{fig:kv_cache_approximation}
\end{figure}

The approach generates a very accurate approximation (Fig.~\ref{fig:kv_cache_approximation}): We evaluated this approach on four greedy MATH-500 \soft rollouts, with an average length of 420 tokens.
We compared caches for the prompt and all generated tokens except the final sampled token, which had not yet been consumed.
Over response positions, three total forward passes achieved a relative $\ell_2$ error of 1.17\% for keys and 1.22\% for values, with 99.88\% next-token argmax agreement. Four passes reduced these errors to 1.08\% and 1.09\%, with little further improvement through 32 passes.
In comparison, we performed one parallel pass using the actual hidden states recorded during sequential decoding, which still produced 0.99\% key error and 0.95\% value error under BF16 execution. The comparison suggests that three to four passes recover the cache close to the numerical discrepancy between parallel and incremental execution on these trajectories.

\subsection{Post-training and likelihood estimation for off-policy rollouts}\label{sec:post_train}

A common post-training procedure samples rollouts from an off-policy sampler and evaluates them under the policy being optimized. The policy gradient then depends on the rollout likelihood, typically weighted by an advantage and an importance ratio. In this section we discuss candidate recipes for off-policy RL.
Note that the solutions discussed are hypothetical only and unverified by experiments.

For on-policy rollouts, likelihood evaluation is essentially free: token log-probabilities can be recorded during decoding.
For an off-policy rollout, however, the latent states under the policy model are not available and must be reconstructed. Under latent-feedback decoding, the input at position $t$ depends on the previous policy hidden state,
\begin{equation}
\bu_t = \be_t \otimes \bh_{t-1}.
\end{equation}
Crucially, $\bh_{t-1}$ cannot be obtained directly from the token prefix alone: it is produced by running position $t-1$ with $\bh_{t-2}$ as latent feedback,
\begin{equation}
\bh_{t-1}
=
f_\theta(\be_{t-1}\otimes \bh_{t-2}; \context_{t-1}).  
\end{equation}
Thus, to evaluate token $t$, the policy must first replay positions $1,\ldots,t-1$ to reconstruct its own latent trajectory. Consequently, denote the rollout as $y_{1:T}$ and the prompt as $x$, we have
\begin{equation}
 \log \pi_\theta(y_{1:T}\mid x)
=
\sum_{t=1}^{T}
\log \pi_\theta(y_t\mid x,y_{<t},h_{t-1}),   
\end{equation}
which cannot be evaluated by standard parallel teacher forcing, but instead requires $O(T)$ sequential steps.

\paragraph{Potential solution 1: Hidden-state replay.}
To avoid sequentially reconstructing the latent trajectory, one can directly reuse the hidden states produced by the behavior policy during sampling, analogous to router replay in MoE post-training~\citep{ma2025stabilizing}.
Specifically, instead of recomputing the policy hidden state $\bh_{t-1}^{\pi}$, we approximate
\[
\bh_{t-1}^{\pi} \approx \bh_{t-1}^{\mu},
\]
where $\mu$ denotes the sampler policy, and evaluate
\[
\log \pi_\theta(y_t \mid y_{<t}, \bh_{t-1}^{\mu}).
\]
This removes the sequential hidden-state replay cost, but introduces bias whenever the latent trajectories of the behavior and policy models differ.

\paragraph{Potential solution 2: Multi-forward pass.} 
Let the exact latent-feedback trajectory satisfy
\begin{equation}
\bh_t^\star = F_{\theta,t}(\be_t,\bh_{t-1}^\star;\context_t),
\end{equation}
which requires sequential replay. Instead, starting from parallel states $\{\bh_t^{(0)}\}$, we perform $K$ Jacobi-style refinement passes,
\[
\bh_t\pass{k}
=
F_{\theta,t}(\be_t,\bh_{t-1}\pass{k-1};\context\pass{k}),
\qquad k=1,\ldots,K,
\]
where all positions are updated in parallel within each pass. Unrolling gives
\begin{equation}
h_t^{(K)}
=
F_{\theta,t}\!\Big(\be_t,
F_{\theta,t-1}\big(\be_{t-1},\ldots,
F_{\theta,t-K+1}(\be_{t-K+1},\bh_{t-K}^{(0)};\context_{t-K+1}\pass{0})\big);\context_t\pass{K}\Big),
\end{equation}
so $K$ passes explicitly recover $K$ hops of the recurrent dependency, while longer-range information is represented only through the initialization $\bh_{t-K}^{(0)}$. We then approximate
\[
\log \pi_\theta(y_{1:T}\mid x)
\approx
\sum_{t=1}^T
\log \pi_\theta
\!\left(y_t\mid x,y_{<t},h_{t-1}^{(K)}\right).
\]
Thus, $K$ controls a finite-horizon approximation to the fully autoregressive latent-feedback distribution, reducing from $O(T)$ sequential steps to $O(K)$ parallel passes over the sequence.

An empirical estimation of the approximation error is shown below in Fig.~\ref{fig:rollout_nll}, where one forward pass denotes just using the embedding as input. The approximation error reduces significantly after two forward passes and plateaus after three passes.

\begin{figure}[!h]
    \centering
    \includegraphics[width=\linewidth]{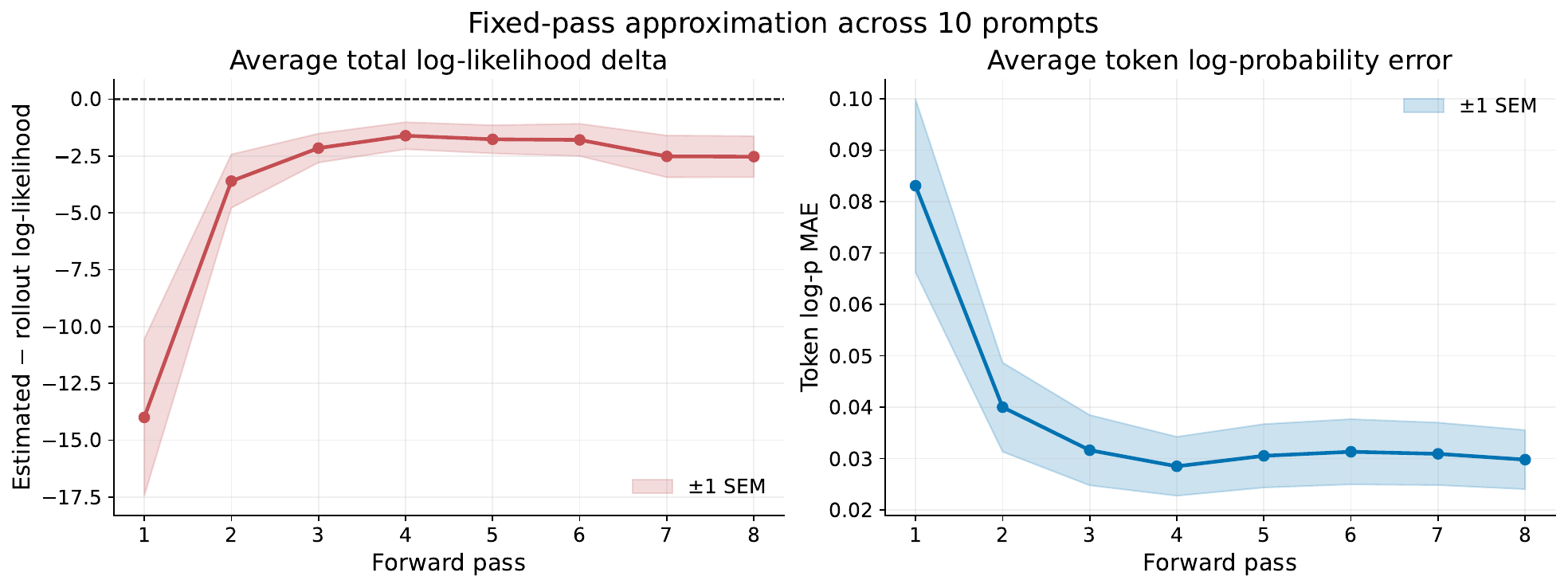}
    \caption{True sequence likelihood can be approximated with multiple forward passes rather than going over all tokens.}
    \label{fig:rollout_nll}
\end{figure}

\paragraph{Gradient truncation through latent feedback}
Given a rollout, the exact score-function gradient contains two paths: a direct contribution through each token prediction, and an indirect contribution through the recurrent hidden states,
\begin{equation}
\nabla_\theta \log \pi_\theta(y_{1:T})
=
\sum_{t=1}^{T}
\left[
\frac{\partial \log \pi_\theta(y_t \mid y_{<t},\bh_{t-1})}{\partial \theta}
+
\frac{\partial \log \pi_\theta(y_t \mid y_{<t},\bh_{t-1})}{\partial \bh_{t-1}}
\frac{d \bh_{t-1}}{d\theta}
\right].
\end{equation}
For the recurrence $\bh_t=F_{\theta,t}(\bh_{t-1})$, the latter term further decomposes through time as
\begin{equation}
\frac{d \bh_{t-1}}{d\theta}
=
\sum_{s=1}^{t-1}
\left(
\prod_{j=s+1}^{t-1}
\frac{\partial \bh_j}{\partial \bh_{j-1}}
\right)
\frac{\partial \bh_s}{\partial \theta}.
\end{equation}
Hence, the gradient of token $t$ contains contributions from all preceding latent-feedback steps. Hidden-state replay removes these terms entirely, since replayed states are produced by the behavior policy and treated as constants. For multi-pass evaluation, detaching $\bh$'s gradient between passes similarly removes gradients through the latent trajectory, whereas backpropagating through $K$ passes retains paths spanning up to $K-1$ latent-feedback hops, analogous to truncated backpropagation through time.

\section{Related work}

\textbf{Latent recurrence across decoding steps.}~
One central idea behind our work is to overlap recurrent compute with the sequential decoding process.
There are other works considering similar ideas. 
\emph{Feedback Transformer}~\citep{fan2020feedback} is the pioneering work along this line; At each position, they generate a mixture of each layer's representation and let attention in future positions attend to the mixture rather than the same-layer key values. However, their training is sequential over input tokens, limiting their scalability, whereas our training is parallelized over all positions. 
Many concurrent works explore a similar direction.
\emph{$T^2$MLR}~\citep{cai2026t}, \emph{Latent Recurrent Transformer}~\citep{huang2026latent}, \emph{Recirculation}~\citep{mozer2026recirculation} consider recurrence with middle-layer injection, where the previous position's hidden state at a late middle layer is fused with the current position's early middle layer hidden state. 
Methodology-wise, our approach is similar in the training approach and the motivation. Our approach mainly differs in the point of reinjection; specifically, our injection happens ``externally'' to the model and therefore introduces no architecture changes. We also introduce the least amount of extra parameters. For a $L$-layer transformer with $D$-dimension residual, we introduce only two linear projections (each of size $D\times D$), in contrast to $T^2$MLR's extra MLPs ($5 D^2$ parameters) and LRT's layerwise projection, which introduces $LD^2$ parameters. 
The bigger difference lies in the scale of empirical evaluation: our work performs the largest-scale pre-training (up to 400B tokens) among all works, enabled by feedback pass scheduling; therefore, we manage to verify the actual \emph{decoding time improvement} on free-form generation tasks, whereas LRT only considers non-freeform eval (similar to our setting in Fig.~\ref{fig:loopy} right), and $T^2$MLR considers synthetic state tracking tasks and GSM8K only after fine-tuning the model on the math corpus. 
However, considering the similarity in spirit, we do not foresee reasons why one approach would outperform the others, and exactly which method (and more broadly, which form of past hidden state injection) works the best \emph{at large scale} remains unclear since we do not have the resources for verification.
Very recent work also explores latent feedback under prefill--decode decoupled architectures, including Maglev \citep{liu2026maglev} and Recurrent Looped Transformer \citep{zhang2026recurrentlooped}, which pairs a context encoder with a recurrent decoder that feeds its output hidden state into the next token's input.

\textbf{Loop transformers.}~
Our approach is similar to loop transformers~\citep{fan2026bridging,dehghani2018universal,giannou2023looped,geiping2025scaling} during training in that the model's outputs are repeatedly fed back as inputs across multiple forward passes.
Differences lie in inference time: loop transformers are recurrent over depth, 
thereby increasing inference compute with the number of reapplying the transformer stack.
In contrast, full-bandwidth transformer performs recurrence over time, integrated into the autoregressive decoding loop, which \emph{avoids additional transformer-block evaluations} per generated token (with additional compute used only when optional multi-pass prefilling is used).


\paragraph{Latent and continuous reasoning.} 
Our approach feeds top layer latent into the context, similar to the central idea of latent reasoning approaches such as Coconut~\citep{hao2024training} and Soft Thinking~\citep{zhang2026soft}. The biggest differences are : (a) We focus on pre-training; (b) We use the hidden state to ``augment'' the generation rather than replacing the discrete tokens, therefore our approach is easier to supervise (but we may be  less token efficient). 
Hybrid Latent Reasoning via Reinforcement Learning~\citep{yue2026hybrid} proposes to use both the top layer hidden state and the generated tokens' embedding at post-training time during rollout, however they did not utilize top layer hidden state but instead they use it to generate a weighted mixture of vocabulary embedding so it is unclear whether it improves the reachability as the full bandwidth transformer does.
There are also works studying latent reasoning at pre-training time, in particular, PonderLM-2~\citep{zeng2025ponderlm} considers an interleaved embedding / hidden state as the input. Notably, their training approach is similar to us in that they use multiple forward passes to replace sequential rollout, however their approach doubles the input length (as well as KV cache size) so they introduce more training and inference overhead than the full bandwidth transformer.

\paragraph{Parallel training of recurrent networks.}
Another related direction is parallel training of recurrent networks.  Most applications of this consider the linear special case like Mamba~\citep{gu2024mamba} or Gated Deltanet~\citep{yang2025gated}.  These are clearly powerful techniques with use in various architectures yet in all such uses they are hybridized with standard transformer layers which can compensate for the missing representational capacity inherited from the linear constraint.
ParaRNN~\citep{danieli2025pararnn} goes further by parallelizing training of nonlinear recurrent neural networks via decoupling the optimizations at each point in the process and using Newton's iterations to achieve convergence with results comparable to transformers for language modeling.  This approach here goes the other way, constructing recurrence on transformers with results that improve over baseline transformers, and it appears that the approach here is significantly more efficient.

\paragraph{Data-efficient pre-training.}
Our work falls into the broad category of improving LLM pre-training's data efficiency, i.e., given the same model size and fixed data, how can we use more flops to build a more powerful model under fixed or more inference overhead.
Existing approaches consider additional objectives (beyond NTP) on the representation~\citep{liu2026next, zhang2026nitp, dai2025context, teoh2025next} that encourage the hidden state to contain richer information.
There has also been a recent NanoGPT slow run competition\footnote{\url{https://qlabs.sh/slowrun/}} that studies this setting, where the official solution~\citep{mandal2026q0} trains a deep ensemble of LLMs and distills them into a single one for better performance.
Compared with these approaches, our framework uses additional training flops for unlocking a new type of decoding regime that gives a free performance boost at inference time.
Additionally, we believe techniques can flow between literature, for example, the depth scaling we used has also been shown to be important for the stability of training loop transformers~\citep{movahedi2026fixed}.
Our empirical verification of recurrence scheduling also suggests the feasibility of introducing computationally intensive auxiliary objectives only later on in the training.

More broadly, these methods point to a shift in the relevant scaling axes for pre-training. Conventional scaling primarily varies model parameters and training tokens. However, in large-scale training, the feasible design space is also constrained by pod size of GPUs, wall-clock budget, and the availability of high-quality unique tokens. Once the token-per-parameter ratio and the accessible pool of high-quality data become binding, simply increasing the number of unique training tokens is no longer the only, or even the most direct, path to improvement. A promising axis is to spend more computation per unique token through recurrent, iterative, or feedback-based mechanisms.

\section{Limitations and future work}\label{sec:limit_and_future}
\paragraph{Larger scale experiments} Our experiment scale is limited to 1B parameter models, and we did not verify the approach on models of larger scale.
However, we believe latent feedback decoding can potentially introduce more benefit for larger ones with more number of layers where the top layer hidden state includes even richer information.
More importantly, the ability for shallow layer to read deep layer information as well as the ability to recursively update a continuous state, is \emph{not} something scaling model size or training horizon can acquire.

\paragraph{More principled scheduling} Current feedback pass scheduling is based on a heuristic; future work can consider more rigorous ablation on the length of the recurrence training phase as well as a more principled approach to determine the number of recurrence steps, e.g., via the Jacobi iteration convergence diagnostics from \citet{zeng2025ponderlm}.

\paragraph{Compatibility with multi-token prediction} Multi-token prediction~\citep[MTP]{gloeckle2024better} generates more than one token at each step at decoding time. However, latent feedback decoding requires a token's previous hidden state to generate the next token; therefore, the current form is incompatible with MTP.
We believe future work can consider further incorporating next latent prediction objective~\citep{teoh2025next} during pre-training, which enables the model to propose multiple future hidden states and use it to predict multiple future tokens with latent feedback decoding.

\paragraph{Turning pre-trained model into full-bandwidth transformer} Our feedback-pass scheduling (Sec.~\ref{sec:recur_training}) effectively turns the full-bandwidth transformer's training into mid-training style, where the majority of the training stays identical as standard pre-training.
Future work can further explore the feasibility of turning publicly available pre-cooldown models, such as the checkpoints from Arcee~\citep{singh2026arcee}\footnote{\url{https://huggingface.co/arcee-ai/Trinity-Nano-Base-Pre-Anneal}} or just pre-trained base models into full-bandwidth transformers.

\paragraph{Post-training} Certain architecture changes require modifications of the post-training algorithm. Particularly, the likelihood evaluation of rollouts under latent feedback now requires a recursive process over all positions for the policy model.
This is fine for on-policy setting, where the hidden states and the likelihood can be captured during decoding. 
However, this can be tricky in off-policy setting, which involves evaluating the likelihood on the policy model position by position. However, one can consider utilizing multi-forward passes with shifted inputs as a multi-gram approximation to the true likelihood. We verify the feasibility in Fig.~\ref{fig:rollout_nll} in the Section \ref{sec:post_train}. 

\paragraph{Interaction with sliding window attention} Our approach has some particular benefit for sliding window attention~\citep[SWA,][]{beltagy2020longformer, jiang2023mistral}. In particular, SWA is known to have small receptive fields.
However, full-bandwidth transformers short-circuit the top-layer hidden state directly to the bottom, which exposes global information to the shallow layers. 
Future work can consider ablating the backbone model's architecture and see whether full-bandwidth transformer enables a higher ratio of SWA for better inference time efficiency and smaller KV cache size.
We provide a more detailed discussion in Appendix~\ref{sec:swa_detail}.

\paragraph{Increased challenge for CoT monitoring} Lastly, latent feedback decoding increased the steps of unverbalized compute between two CoT~\citep{brown2026quantifying, engels2026transparent} steps, therefore increasing the difficulty of CoT monitoring~\citep{baker2025monitoring}. Future work can consider building probing based monitoring tools that directly reads the hidden states rather than verbal CoT. The issue is also recently discussed in the community~\footnote{\url{https://www.redwoodresearch.org/blog/an-operationalization-of-opaque-serial-depth}}.

\section*{Discussion: Origins of the idea.} 
The initial motivation for this work came from observations in NextLat~\citep{teoh2025next} where predicting latent states yields a sequence of fixed points, suggesting the feasibility of parallel training a recursive model.

This led us to ask whether a transformer could be trained recursively in parallel by feeding back soft latent belief states across multiple passes. The key shift was to view transformers not merely as fixed-depth feedforward transformations, but as transition operators acting over latent belief states, enabling recursive computation without losing parallelizability.

The authors would also like to thank Jayden Teoh for discussion on an earlier version of the idea, which turned into experiments in the NextLat paper on $NC^1$-hard tasks, where the author observed that while the transformer failed to generalize beyond the trained sequence length, the co-trained latent dynamics model was able to roll out without error during inference and generalize to all lengths.





\bibliography{refs}

@inproceedings{gu2024mamba,
  title={Mamba: Linear-time sequence modeling with selective state spaces},
  author={Gu, Albert and Dao, Tri},
  booktitle={First conference on language modeling},
  year={2024}
}

@inproceedings{yang2025gated,
  title={Gated delta networks: Improving mamba2 with delta rule},
  author={Yang, Songlin and Kautz, Jan and Hatamizadeh, Ali},
  booktitle={International Conference on Learning Representations},
  volume={2025},
  pages={29687--29707},
  year={2025}
}

@article{danieli2025pararnn,
  title={Pararnn: Unlocking parallel training of nonlinear rnns for large language models},
  author={Danieli, Federico and Rodriguez, Pau and Sarabia, Miguel and Suau, Xavier and Zappella, Luca},
  journal={arXiv preprint arXiv:2510.21450},
  year={2025}
}

@article{fan2020feedback,
  title={Addressing Some Limitations of Transformers with Feedback Memory},
  author={Fan, Angela and Lavril, Thibaut and Grave, Edouard and Joulin, Armand and Sukhbaatar, Sainbayar},
  journal={arXiv preprint arXiv:2002.09402},
  year={2020}
}

@article{liu2026next,
  title={Next Concept Prediction in Discrete Latent Space Leads to Stronger Language Models},
  author={Liu, Yuliang and Song, Yunchong and Wang, Yixuan and Ge, Kewen and Lamb, Alex and Guo, Qipeng and Chen, Kai and Zhou, Bowen and Lin, Zhouhan},
  journal={arXiv preprint arXiv:2602.08984},
  year={2026}
}

@inproceedings{zhang2026nitp,
  title={NITP: Next Implicit Token Prediction for LLM Pre-training},
  author={Zhang, Xiangdong and Zhang, Debing and Zhang, Shaofeng and Qin, Xiaohan and Cheng, Yu and Yan, Junchi},
  booktitle={Forty-third International Conference on Machine Learning},
  year={2026}
}

@article{mandal2026q0,
  title={q0: Primitives for Hyper-Epoch Pretraining},
  author={Mandal, Bishwas and Berman, Shmuel and Vegesna, Akshay and Dahal, Samip},
  journal={arXiv preprint arXiv:2606.03938},
  year={2026}
}

@article{dai2025context,
  title={Context-level language modeling by learning predictive context embeddings},
  author={Dai, Beiya and Liu, Yuliang and Xue, Daozheng and Song, Yunchong and Guo, Qipeng and Chen, Kai and Wang, Xinbing and Zhou, Bowen and Lin, Zhouhan},
  journal={arXiv preprint arXiv:2510.20280},
  year={2025}
}

@article{li2024eagle,
  title={Eagle: Speculative sampling requires rethinking feature uncertainty},
  author={Li, Yuhui and Wei, Fangyun and Zhang, Chao and Zhang, Hongyang},
  journal={arXiv preprint arXiv:2401.15077},
  year={2024}
}

@article{gloeckle2024better,
  title={Better \& faster large language models via multi-token prediction},
  author={Gloeckle, Fabian and Idrissi, Badr Youbi and Rozi{\`e}re, Baptiste and Lopez-Paz, David and Synnaeve, Gabriel},
  journal={arXiv preprint arXiv:2404.19737},
  year={2024}
}

@inproceedings{
li2026normuon,
title={NorMuon: Making Muon more efficient and scalable},
author={Zichong Li and Liming Liu and Chen Liang and Weizhu Chen and Tuo Zhao},
booktitle={Forty-third International Conference on Machine Learning},
year={2026},
url={https://openreview.net/forum?id=m1IRWFAMsa}
}

@inproceedings{li2024chain,
  title={Chain of thought empowers transformers to solve inherently serial problems},
  author={Li, Zhiyuan and Liu, Hong and Zhou, Denny and Ma, Tengyu},
  booktitle={International Conference on Learning Representations},
  volume={2024},
  pages={11911--11943},
  year={2024}
}

@article{cai2026t,
  title={T\^{} 2MLR: Transformer with Temporal Middle-Layer Recurrence},
  author={Cai, Ziyang and Zhu, Xingyu and Dong, Yihe and He, Yinghui and Arora, Sanjeev},
  journal={arXiv preprint arXiv:2607.15178},
  year={2026}
}

@article{huang2026latent,
  title={Latent Recurrent Transformer: Architecture Exploration, Training Strategies, and Scaling Behavior},
  author={Huang, Zeyi and He, Xuehai and Ren, LiLiang and Wang, Yiping and Peng, Baolin and Cheng, Hao and Wang, Shuohang and He, Pengcheng and Gao, Jianfeng and Lee, Yong Jae and others},
  journal={arXiv preprint arXiv:2605.26797},
  year={2026}
}

@article{movahedi2026fixed,
  title={Fixed-Point Reasoners: Stable and Adaptive Deep Looped Transformers},
  author={Movahedi, Sajad and Milovanovi{\'c}, Vera and Feigin, Shlomo Libo and Theus, Alexander and Hofmann, Thomas and Boeva, Valentina and Rusch, T Konstantin and Orvieto, Antonio},
  journal={arXiv preprint arXiv:2606.18206},
  year={2026}
}

@article{zeng2025ponderlm,
  title={Ponderlm-2: Pretraining llm with latent thoughts in continuous space},
  author={Zeng, Boyi and Li, He and Song, Shixiang and Wang, Yixuan and Wang, Zitong and He, Ziwei and Wang, Xinbing and Lin, Zhouhan},
  journal={arXiv preprint arXiv:2509.23184},
  year={2025}
}

@article{fan2026bridging,
  title={Bridging the Gap Between Latent and Explicit Reasoning with Looped Transformers},
  author={Fan, Ying and Svete, Anej and Lee, Kangwook},
  journal={arXiv preprint arXiv:2606.31779},
  year={2026}
}

@article{dehghani2018universal,
  title={Universal transformers},
  author={Dehghani, Mostafa and Gouws, Stephan and Vinyals, Oriol and Uszkoreit, Jakob and Kaiser, {\L}ukasz},
  journal={arXiv preprint arXiv:1807.03819},
  year={2018}
}

@inproceedings{giannou2023looped,
  title={Looped transformers as programmable computers},
  author={Giannou, Angeliki and Rajput, Shashank and Sohn, Jy-yong and Lee, Kangwook and Lee, Jason D and Papailiopoulos, Dimitris},
  booktitle={International Conference on Machine Learning},
  pages={11398--11442},
  year={2023},
  organization={PMLR}
}

@article{hao2024training,
  title={Training large language models to reason in a continuous latent space},
  author={Hao, Shibo and Sukhbaatar, Sainbayar and Su, DiJia and Li, Xian and Hu, Zhiting and Weston, Jason and Tian, Yuandong},
  journal={arXiv preprint arXiv:2412.06769},
  year={2024}
}

@article{yue2026hybrid,
  title={Hybrid latent reasoning via reinforcement learning},
  author={Yue, Zhenrui and Jin, Bowen and Zeng, Huimin and Zhuang, Honglei and Qin, Zhen and Yoon, Jinsung and Shang, Lanyu and Han, Jiawei and Wang, Dong},
  journal={Advances in Neural Information Processing Systems},
  volume={38},
  pages={5501--5530},
  year={2026}
}

@article{zhang2026soft,
  title={Soft thinking: Unlocking the reasoning potential of llms in continuous concept space},
  author={Zhang, Zhen and He, Xuehai and Yan, Weixiang and Shen, Ao and Zhao, Chenyang and Wang, Xin},
  journal={Advances in Neural Information Processing Systems},
  volume={38},
  pages={168990--169012},
  year={2026}
}

@article{cobbe2021training,
  title={Training verifiers to solve math word problems},
  author={Cobbe, Karl and Kosaraju, Vineet and Bavarian, Mohammad and Chen, Mark and Jun, Heewoo and Kaiser, Lukasz and Plappert, Matthias and Tworek, Jerry and Hilton, Jacob and Nakano, Reiichiro and others},
  journal={arXiv preprint arXiv:2110.14168},
  year={2021}
}

@article{lightman2023lets,
      title={Let's Verify Step by Step}, 
      author={Lightman, Hunter and Kosaraju, Vineet and Burda, Yura and Edwards, Harri and Baker, Bowen and Lee, Teddy and Leike, Jan and Schulman, John and Sutskever, Ilya and Cobbe, Karl},
      journal={arXiv preprint arXiv:2305.20050},
      year={2023}
}

@article{chen2021codex,
  title={Evaluating Large Language Models Trained on Code},
  author={Mark Chen and Jerry Tworek and Heewoo Jun and Qiming Yuan and Henrique Ponde de Oliveira Pinto and Jared Kaplan and Harri Edwards and Yuri Burda and Nicholas Joseph and Greg Brockman and Alex Ray and Raul Puri and Gretchen Krueger and Michael Petrov and Heidy Khlaaf and Girish Sastry and Pamela Mishkin and Brooke Chan and Scott Gray and Nick Ryder and Mikhail Pavlov and Alethea Power and Lukasz Kaiser and Mohammad Bavarian and Clemens Winter and Philippe Tillet and Felipe Petroski Such and Dave Cummings and Matthias Plappert and Fotios Chantzis and Elizabeth Barnes and Ariel Herbert-Voss and William Hebgen Guss and Alex Nichol and Alex Paino and Nikolas Tezak and Jie Tang and Igor Babuschkin and Suchir Balaji and Shantanu Jain and William Saunders and Christopher Hesse and Andrew N. Carr and Jan Leike and Josh Achiam and Vedant Misra and Evan Morikawa and Alec Radford and Matthew Knight and Miles Brundage and Mira Murati and Katie Mayer and Peter Welinder and Bob McGrew and Dario Amodei and Sam McCandlish and Ilya Sutskever and Wojciech Zaremba},
  year={2021},
  eprint={2107.03374},
  archivePrefix={arXiv},
  primaryClass={cs.LG}
}

@article{austin2021program,
  title={Program Synthesis with Large Language Models},
  author={Austin, Jacob and Odena, Augustus and Nye, Maxwell and Bosma, Maarten and Michalewski, Henryk and Dohan, David and Jiang, Ellen and Cai, Carrie and Terry, Michael and Le, Quoc and others},
  journal={arXiv preprint arXiv:2108.07732},
  year={2021}
}

@article{teoh2025next,
  title={Next-latent prediction transformers learn compact world models},
  author={Teoh, Jayden and Tomar, Manan and Ahn, Kwangjun and Hu, Edward S and Pearce, Tim and Sharma, Pratyusha and Krishnamurthy, Akshay and Islam, Riashat and Lamb, Alex and Langford, John},
  journal={arXiv preprint arXiv:2511.05963},
  year={2025}
}

@inproceedings{
yang2024tensor,
title={Tensor Programs {VI}: Feature Learning in Infinite Depth Neural Networks},
author={Greg Yang and Dingli Yu and Chen Zhu and Soufiane Hayou},
booktitle={The Twelfth International Conference on Learning Representations},
year={2024},
url={https://openreview.net/forum?id=17pVDnpwwl}
}

@article{noci2022signal,
  title={Signal propagation in transformers: Theoretical perspectives and the role of rank collapse},
  author={Noci, Lorenzo and Anagnostidis, Sotiris and Biggio, Luca and Orvieto, Antonio and Singh, Sidak Pal and Lucchi, Aurelien},
  journal={Advances in Neural Information Processing Systems},
  volume={35},
  pages={27198--27211},
  year={2022}
}

@article{defazio2025gradients,
  title={Why gradients rapidly increase near the end of training},
  author={Defazio, Aaron},
  journal={arXiv preprint arXiv:2506.02285},
  year={2025}
}

@article{hagele2024scaling,
  title={Scaling laws and compute-optimal training beyond fixed training durations},
  author={H{\"a}gele, Alexander and Bakouch, Elie and Kosson, Atli and Allal, Loubna B and Von Werra, Leandro and Jaggi, Martin},
  journal={Advances in Neural Information Processing Systems},
  volume={37},
  pages={76232--76264},
  year={2024}
}

@article{hu2024minicpm,
  title={Minicpm: Unveiling the potential of small language models with scalable training strategies},
  author={Hu, Shengding and Tu, Yuge and Han, Xu and He, Chaoqun and Cui, Ganqu and Long, Xiang and Zheng, Zhi and Fang, Yewei and Huang, Yuxiang and Zhao, Weilin and others},
  journal={arXiv preprint arXiv:2404.06395},
  year={2024}
}

@article{chowdhery2023palm,
  title={Palm: Scaling language modeling with pathways},
  author={Chowdhery, Aakanksha and Narang, Sharan and Devlin, Jacob and Bosma, Maarten and Mishra, Gaurav and Roberts, Adam and Barham, Paul and Chung, Hyung Won and Sutton, Charles and Gehrmann, Sebastian and others},
  journal={Journal of machine learning research},
  volume={24},
  number={240},
  pages={1--113},
  year={2023}
}

@article{abdin2024phi,
  title={Phi-4 technical report},
  author={Abdin, Marah and Aneja, Jyoti and Behl, Harkirat and Bubeck, S{\'e}bastien and Eldan, Ronen and Gunasekar, Suriya and Harrison, Michael and Hewett, Russell J and Javaheripi, Mojan and Kauffmann, Piero and others},
  journal={arXiv preprint arXiv:2412.08905},
  year={2024}
}

@article{qi2025evolm,
  title={EvoLM: In search of lost language model training dynamics},
  author={Qi, Zhenting and Nie, Fan and Alahi, Alexandre and Zou, James and Lakkaraju, Himabindu and Du, Yilun and Xing, Eric and Kakade, Sham and Zhang, Hanlin},
  journal={arXiv preprint arXiv:2506.16029},
  year={2025}
}

@article{wei2022chain,
  title={Chain-of-thought prompting elicits reasoning in large language models},
  author={Wei, Jason and Wang, Xuezhi and Schuurmans, Dale and Bosma, Maarten and Xia, Fei and Chi, Ed and Le, Quoc V and Zhou, Denny and others},
  journal={Advances in neural information processing systems},
  volume={35},
  pages={24824--24837},
  year={2022}
}

@inproceedings{
geiping2025scaling,
title={Scaling up Test-Time Compute with Latent Reasoning: A Recurrent Depth Approach},
author={Jonas Geiping and Sean Michael McLeish and Neel Jain and John Kirchenbauer and Siddharth Singh and Brian R. Bartoldson and Bhavya Kailkhura and Abhinav Bhatele and Tom Goldstein},
booktitle={The Thirty-ninth Annual Conference on Neural Information Processing Systems},
year={2025},
url={https://openreview.net/forum?id=S3GhJooWIC}
}

@article{singh2026arcee,
  title={Arcee trinity large technical report},
  author={Singh, Varun and Krauss, Lucas and Jaghouar, Sami and Sirovatka, Matej and Goddard, Charles and Obied, Fares and Ong, Jack Min and Straube, Jannik and Harley, Aria and Stewart, Conner and others},
  journal={arXiv preprint arXiv:2602.17004},
  year={2026}
}

@article{sharma2026dense,
  title={Dense Supervision Is Not Enough: The Readout Blind Spot in Looped Language Models},
  author={Sharma, Rituraj and Vu, Tu},
  journal={arXiv preprint arXiv:2606.24898},
  year={2026}
}

@article{shomali2026loopmtp,
  title={LoopMTP: A looped transformer guided by latent multi-token prediction},
  author={Shomali, Behzad and Frey, Markus and Berghaus, David and Koehler, Joachim and Ali, Mehdi},
  journal={arXiv preprint arXiv:2608.03624},
  year={2026}
}

@article{zhu2025scaling,
  title={Scaling latent reasoning via looped language models},
  author={Zhu, Rui-Jie and Wang, Zixuan and Hua, Kai and Zhang, Tianyu and Li, Ziniu and Que, Haoran and Wei, Boyi and Wen, Zixin and Yin, Fan and Xing, He and others},
  journal={arXiv preprint arXiv:2510.25741},
  year={2025}
}

@article{beltagy2020longformer,
  title={Longformer: The long-document transformer},
  author={Beltagy, Iz and Peters, Matthew E and Cohan, Arman},
  journal={arXiv preprint arXiv:2004.05150},
  year={2020}
}

@article{jiang2023mistral,
  title={Mistral 7b. arxiv},
  author={Jiang, Albert Q and Sablayrolles, A and Mensch, A and Bamford, C and Chaplot, D Singh and Casas, Ddl and Bressand, F and Lengyel, G and Lample, G and Saulnier, L and others},
  journal={arXiv preprint arXiv:2310.06825},
  volume={10},
  number={3},
  year={2023}
}

@article{engels2026transparent,
  title={How Transparent is DiffusionGemma?},
  author={Engels, Joshua and McDougall, Callum and Chughtai, Bilal and Kramar, Janos and Rajamanoharan, Senthoran and Wu, Cindy and Conmy, Arthur and Chen, Asic Q and Tarbouriech, Jean and Ma, Min and others},
  journal={arXiv preprint arXiv:2606.20560},
  year={2026}
}

@article{brown2026quantifying,
  title={Quantifying the Necessity of Chain of Thought through Opaque Serial Depth},
  author={Brown-Cohen, Jonah and Lindner, David and Shah, Rohin},
  journal={arXiv preprint arXiv:2603.09786},
  year={2026}
}

@article{baker2025monitoring,
  title={Monitoring reasoning models for misbehavior and the risks of promoting obfuscation},
  author={Baker, Bowen and Huizinga, Joost and Gao, Leo and Dou, Zehao and Guan, Melody Y and Madry, Aleksander and Zaremba, Wojciech and Pachocki, Jakub and Farhi, David},
  journal={arXiv preprint arXiv:2503.11926},
  year={2025}
}

@article{liu2026maglev,
  title={Maglev: Sliding Recurrent Memory},
  author={Liu, Bo and Liu, Qiang},
  journal={arXiv preprint arXiv:2608.02870},
  year={2026}
}

@article{zhou2023instruction,
  title={Instruction-following evaluation for large language models},
  author={Zhou, Jeffrey and Lu, Tianjian and Mishra, Swaroop and Brahma, Siddhartha and Basu, Sujoy and Luan, Yi and Zhou, Denny and Hou, Le},
  journal={arXiv preprint arXiv:2311.07911},
  year={2023}
}

@article{wu2026writingbench,
  title={Writingbench: A comprehensive benchmark for generative writing},
  author={Wu, Yuning and Mei, Jiahao and Yan, Ming and Li, Chenliang and Lai, Shaopeng and Ren, Yuran and Wang, Zijia and Zhang, Ji and Wu, Mengyue and Jin, Qin and others},
  journal={Advances in Neural Information Processing Systems},
  volume={38},
  year={2026}
}

@article{mozer2026recirculation,
  title={Recirculation},
  author={Mozer, Michael C and Siddiqui, Shoaib Ahmed and Sawyer, Danny and Sanyal, Sunny and Liu, Rosanne},
  journal={arXiv preprint arXiv:2608.17981},
  year={2026}
}

@techreport{zhang2026recurrentlooped,
  title  = {Recurrent Looped Transformer},
  author = {Zhang, Yifan},
  year   = {2026},
  month  = sep,
  url    = {https://github.com/yifanzhang-pro/recurrent-looped-tranformer}
}

@article{ma2026mopd,
  title={Mopd: Multi-teacher on-policy distillation for capability integration in llm post-training},
  author={Ma, Wenhan and Wei, Jianyu and Zhao, Liang and Zhang, Hailin and Xiao, Bangjun and Li, Lei and Yang, Qibin and Gao, Bofei and Wang, Yudong and Li, Rang and others},
  journal={arXiv preprint arXiv:2606.30406},
  year={2026}
}

@article{ma2025stabilizing,
  title={Stabilizing moe reinforcement learning by aligning training and inference routers},
  author={Ma, Wenhan and Zhang, Hailin and Zhao, Liang and Song, Yifan and Wang, Yudong and Sui, Zhifang and Luo, Fuli},
  journal={arXiv preprint arXiv:2510.11370},
  year={2025}
}
\newpage

\appendix

\section{Model architecture}\label{sec:model_arch}

The model is a decoder-only causal language model with a tied 100,352-token embedding and output head, 24 transformer layers, a 1,536-dimensional hidden state, and 6,656-dimensional SiLU GLU feed-forward blocks. Its gated grouped-query attention uses 16 query heads, 8 shared key/value heads, headwise gates, QK RMS normalization, and rotary positions over an 8,192-token context; most layers use a 2,048-token sliding window, while every sixth layer uses full attention. RMS normalization is applied around each residual block and at the final output.

\section{vLLM compatibility}\label{sec:vllm}
The implementation on vLLM follows the same design pattern as EAGLE~\citet{li2024eagle} / MTP~\citet{gloeckle2024better}:
it retains each request's latest trunk hidden state and copies it in place into a persistent, fixed-address model buffer before the next decode step, allowing CUDA graphs to capture the glu cross gate (Eq.~\eqref{eq:glu_cross}) inside \texttt{forward}.
A patched \texttt{GPUModelRunner.\_model\_forward} stores detached hidden states in a dictionary keyed by request ID, uses \texttt{query\_start\_loc} to map packed rows to requests, and removes completed requests. 
Our forward function then fuses the saved state with the next token embedding through the learned glu cross gate, then recycles the resulting hidden state.
Unlike EAGLE/MTP, which send target hidden states to a separate speculative draft model, our model feeds its own state back into the same model to define the actual next-token distribution. 

\section{Full pseudo code for training and inference}

\begin{figure}[!h]
\begin{minipage}[t]{\textwidth}
\begin{lstlisting}[style=lfd, caption={Training: one step with $k$ passes.}]
def glu_cross(h, e):      # [T,D],[T,D]->[T,D]
    return (h @ W_u) * sigmoid(e @ W_g)
 
e = embed(tokens)         # [T, D]
h = model(e)              # pass 1 (standard)
loss = ntp_loss(h)
for _ in range(k - 1):    # parallel in T
    h = h + uniform(-delta, delta) # jitter noise
    x = glu_cross(shift_right(h), input_rmsnorm_1(e))
    x = prefix_mixin(x, e) # random plain prefix
    h = model(input_rmsnorm_2(x))
    loss += ntp_loss(h) / (k-1)
\end{lstlisting}

\begin{lstlisting}[style=lfd, caption={Inference (\soft); uncommenting line 2-3 gives \fused; line 7 shows the \std-decoding input).}]
h = model(embed(prompt)) # prefill, h: [T, D]
#h = model(
#   glu_cross(shift_right(h), embed(prompt)))
tok = sample(lm_head(h[-1]))
h_prev = h[-1]
while not done:             # decode
    x = glu_cross(h_prev, input_rmsnorm_1(embed(tok)))
    # standard decoding: x = embed(tok)
    h_prev = model_step(input_rmsnorm_2(x), kv_cache)
    tok = sample(lm_head(h_prev))
\end{lstlisting}
\end{minipage}
\caption{Full training and inference code for full-bandwidth transformer, with normalization layer and regularization noise included.
The jitter noise is not applied at inference time but the two rmsnorm still do.
}
\label{fig:pseudocode_complete}
\end{figure}

\section{End-to-end per step training time}\label{sec:train_runtime}
Independent runs tested on 8 B200 GPUs in internal training infrastructure, with gradient checkpointing used to avoid OOM. Each additional pass adds approximately 0.4 seconds extra per iteration.
\begin{itemize}
    \item 1 pass (standard ntp): 0.36 seconds
    \item 2 pass: 0.75 seconds
    \item 3 pass: 1.13 seconds
    \item 4 pass: 1.53 seconds
\end{itemize}

\newpage
\section{Feedback pass scheduling ablation}\label{sec:schedule_ablation}
\begin{figure}[!h]
\begin{subfigure}[t]{.45\linewidth}
        \includegraphics[width=\linewidth]{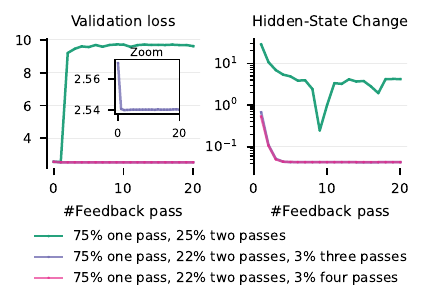}\caption{200B tokens}
        \label{fig:200b_extra_extended}
    \end{subfigure}~~
\begin{subfigure}[t]{.45\linewidth}
        \includegraphics[width=\linewidth]{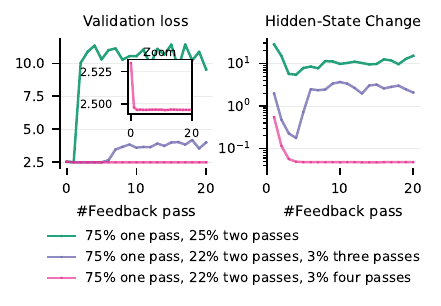}\caption{400B tokens}
        \label{fig:400b_extra_extended}
    \end{subfigure}~~
    
\centering 
\begin{subfigure}[t]{.45\linewidth}
    \includegraphics[width=\linewidth]{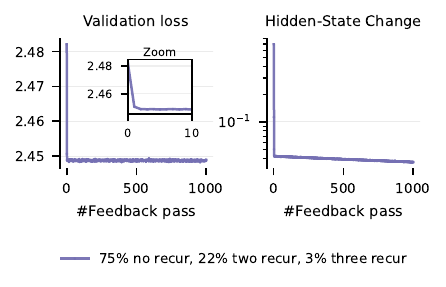}
    \caption{
    Similar to Fig.~\ref{fig:length_extrapolation} and \ref{fig:200b_extra_extended}, but extending number of feedback pass to 1,000.
    The extrapolation remains stable far beyond the 3 passes used in training time. 
    }
    \label{fig:length_extrapolation_100}
\end{subfigure}
\begin{subfigure}{.45\linewidth}
\includegraphics[width=\linewidth]{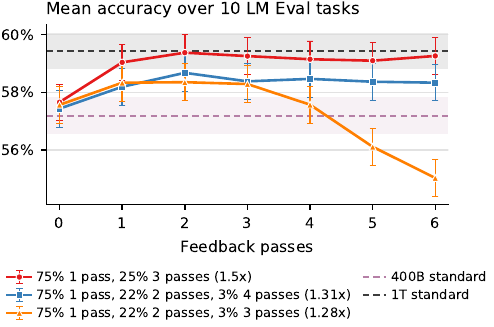}
\caption{LM Eval results, 400B tokens scale}\label{fig:lm_eval_400b_extra}
\end{subfigure}

\begin{subfigure}[t]{.95\linewidth}
    \includegraphics[width=\linewidth]{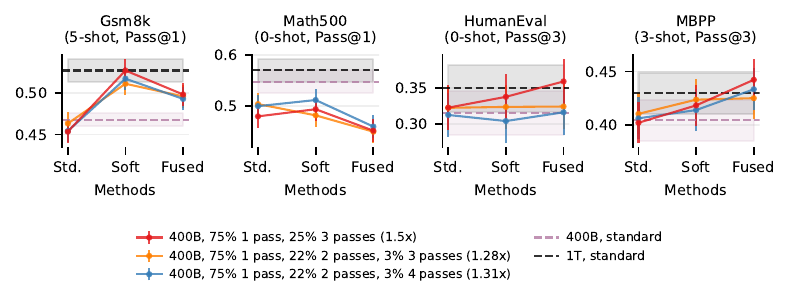}\caption{Free form generation results, 400B tokens scale}
    \label{fig:free_form_400b_extra}
\end{subfigure}~~

\end{figure}

The schedule of feedback passes is based on heuristics, and pursuing extreme efficiency can result in instability. For instance, the 75\% one pass, 22\% two passes and 3\% three passes schedule we suggested in the main text enables a stationary point when trained on 200B tokens (blue line, Fig.\ref{fig:200b_extra_extended}) but failed to extrapolate at 400B tokens scale (blue line, Fig.~\ref{fig:400b_extra_extended} and orange line, Fig.~\ref{fig:lm_eval_400b_extra}).
However, changing the last stage from three to four passes or having three passes across the last 25\% fixes the instability (purple line, Fig.~\ref{fig:400b_extra_extended}; blue and red lines, Fig.~\ref{fig:lm_eval_400b_extra}).


\newpage
\section{Comparison of LM eval performance with other models of similar scale}\label{sec:lm_eval_comp}

\begin{table}[!h]
\small
\centering
\setlength{\tabcolsep}{6pt}
\renewcommand{\arraystretch}{1.1}
\begin{tabular}{llcccccc}
\specialrule{1.5pt}{0pt}{0pt}
\textbf{Model Name} & \textbf{Tokens} & \textbf{W/G} & \textbf{PIQA} & \textbf{OBQA} & \textbf{ARC-E} & \textbf{ARC-C} & \textbf{Avg.} \\
\midrule
OPT 1.3B     & 300B & 59.59 & 72.36 & 33.40 & 50.80 & 29.44 & \textcolor{red}{49.87} \\
Pythia 1B    & 300B & 53.43 & 69.21 & 31.40 & 48.99 & 27.05 & \textcolor{red}{46.21} \\
Pythia 1.4B  & 300B & 57.38 & 70.95 & 33.20 & 54.00 & 28.50 & \textcolor{red}{49.34} \\
TinyLlama 1B & 2T   & 59.43 & 73.56 & 36.80 & 55.47 & 32.68 & \textcolor{green}{53.23} \\
Llama3.2 1B  & 9T   & 60.46 & 74.54 & 37.00 & 60.48 & 35.75 & 55.31 \\
Qwen3 1.7B   & 36T  & 61.01 & 72.36 & 36.80 & 69.91 & 43.26 & 57.30 \\
\midrule
\multirow{5}{*}{\makecell[l]{EvoLM 1B\\ \citep{qi2025evolm}}} & 20B  & 51.30 & 67.85 & 32.80 & 54.80 & 29.61 & 46.44 \\
                                          & 40B  & 54.62 & 69.59 & 36.20 & 58.08 & 30.29 & 49.38 \\
                                          & 80B  & 53.59 & 70.78 & 37.20 & 62.71 & 35.92 & 51.88 \\
                                          & 160B & 53.99 & 71.71 & 36.60 & 63.09 & 36.09 & 52.30 \\
                                          & 320B & 53.51 & 71.93 & 37.20 & 62.29 & 36.18 & 52.49 \\
\midrule
\multirow{2}{*}{\makecell[l]{Full-bandwidth\\transformer 1B}} & 200B (0 feedback pass) & 60.46 & 71.11 & 34.60 & 62.42 & 34.73 & 52.66 \\
                                          & 200B (1 feedback pass)     & 62.59 & 71.49 & 35.00 & 63.43 & 35.41 & \textbf{53.58} \\
\specialrule{1.5pt}{0pt}{0pt}
\end{tabular}
\caption{0-shot LM Eval performance comparison, numbers for EvoLM and other open-sourced models are adopted from Table 4 in the appendix of \citet{qi2025evolm}.}
\label{tab:obs-comparison}
\end{table}


\section{Additional evaluation results for instruction-tuned model}

We additionally evaluate the model on instruction following benchmark, IFEval~\citep{zhou2023instruction} and the English subset of WritingBench~\citep {wu2026writingbench}, using the official fine-tuned Qwen 2.5 7B as the critic model.

For IFEval, we use greedy decoding, and for WritingBench, we use the official suggested random decoding setting, with temperature 0.7, top-p of 0.8 and top-k of 20.

The results are shown in table \ref{tab:if_results_extended}. Broadly, we did not see full-bandwidth models differ from standard model in these two evaluation settings.


\begin{table}[h!]
\centering
\small
\setlength{\tabcolsep}{3.2pt}
\begin{tabular}{lccccc}
\toprule
& \multicolumn{2}{c}{Standard Transformer}
& \multicolumn{3}{c}{Full-bandwidth, 400B} \\
\cmidrule(lr){2-3}
\cmidrule(lr){4-6}
Task
& 400B & 1T
& \stds & \soft & \fused \\
\midrule
WritingBench
& 42.0 & 42.7 & 42.04 & 42.51 & 42.90 \\
\midrule
IFEval (prompt strict)
& 34.94 & 36.60 & 35.30 & 34.57 & 34.75 \\
IFEval (instruction strict)
& 45.44 & 45.92 & 46.76 & 45.08 & 44.60 \\
\bottomrule
\end{tabular}
\caption{Some other results.}
\label{tab:if_results_extended}
\end{table}


\newpage
\section{Concise reasoning}
Fig.~\ref{fig:rollout_length_and_acc} in the main text demonstrates the shortened rollout length for the 200B tokens model. Same observations on \soft and \fused shortening reasoning than \std also hold on other 400B tokens scale run under the three different schedule considered in Sec.~\ref{sec:schedule_ablation}, shown in Fig.~\ref{fig:rollout_length_and_acc_extended} below.

\begin{figure}[!h]
    \centering
    \begin{subfigure}[t]{0.45\linewidth}
        \centering
        \includegraphics[width=\linewidth]{figures/reasoning_length_medium_level.pdf}
        \caption{200B, same as Fig.~\ref{fig:rollout_length_and_acc}.}
    \end{subfigure}
    \hfill
    \begin{subfigure}[t]{0.45\linewidth}
        \centering
        \includegraphics[width=\linewidth]{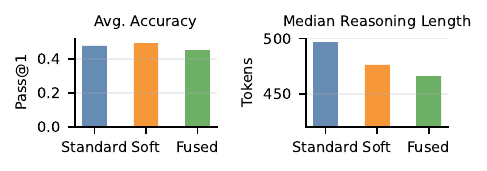}
        \caption{400B, $1.5\times$.}
    \end{subfigure}

    \vspace{0.5em}

    \begin{subfigure}[t]{0.45\linewidth}
        \centering
        \includegraphics[width=\linewidth]{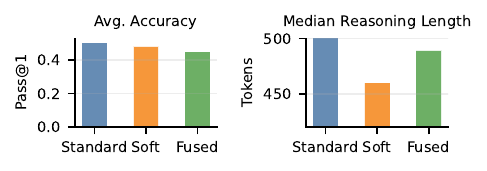}
        \caption{400B, $1.28\times$.}
    \end{subfigure}
    \hfill
    \begin{subfigure}[t]{0.45\linewidth}
        \centering
        \includegraphics[width=\linewidth]{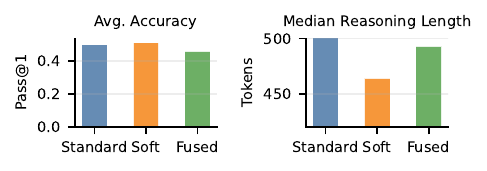}
        \caption{400B, $1.3\times$.}
    \end{subfigure}

    \caption{
        Reasoning length and accuracy on 0-shot MATH-500 across different runs.
    }
    \label{fig:rollout_length_and_acc_extended}
\end{figure}

\section{Generation length}
In the table below, we summarize the median generation length, aggregated over all samples in the benchmark dataset. Note that the length is far beyond the three-token horizon the full bandwidth model is exposed to during training, confirming that at inference time, the model manages to extrapolate a recurrent horizon way longer than training time.
\begin{table}[htbp]
\centering
\begin{tabular}{lrccc}
\toprule
\textbf{task} & \textbf{budget} & \textbf{\std (mean/med)} & \textbf{\soft (mean/med)} & \textbf{\fused (mean/med)} \\
\midrule
gsm8k        & 512  & 242.1 / 231 & 234.5 / 224   & 239.9 / 233 \\
math500      & 2048 & 619.5 / 486 & 607.2 / 467   & 600.4 / 471 \\
humaneval    & 1024 & 131.7 / 96  & 137.8 / 108   & 143.8 / 119 \\
mbpp         & 1024 & 84.3 / 42   & 78.4 / 43     & 72.2 / 44   \\
ifeval       & 1536 & 400.0 / 279 & 375.4 / 256   & 376.3 / 256 \\
WritingBench & 4096 & 688.8 / 690 & 643.2 / 679  & 643.2 / 679 \\
\bottomrule
\end{tabular}
\caption{
Generation length statistics for each task (mean / median tokens). Top 4 tasks are based on base model whereas the last two tasks are based on instruction tuned model.
}\label{tab:iycj_generation_length}
\end{table}

\newpage
\section{Number of pass v.s. free form results}
For the 400B token $1.5\times$ run in table.~\ref{tab:if_results}, we evaluated \fused under additional prefill pass. The results are shown below, where we do not observe further improvement. Notably, the performance does not peak at the number of passes used at training time.

\begin{figure}[!h]
    \centering
    \includegraphics[width=\linewidth]{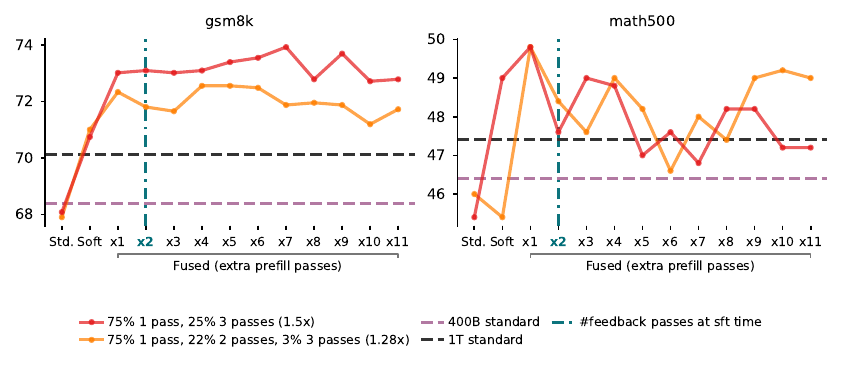}
    \caption{\textbf{Increasing the number of prefill pass does not further improve \fused's performance.}}
    \label{fig:num_pass_ablation}
\end{figure}

\newpage
\section{Interaction of full-bandwidth transformer with sliding window attention}
\label{sec:swa_detail}

\begin{figure}[!h]
    \centering
    \includegraphics[width=\linewidth]{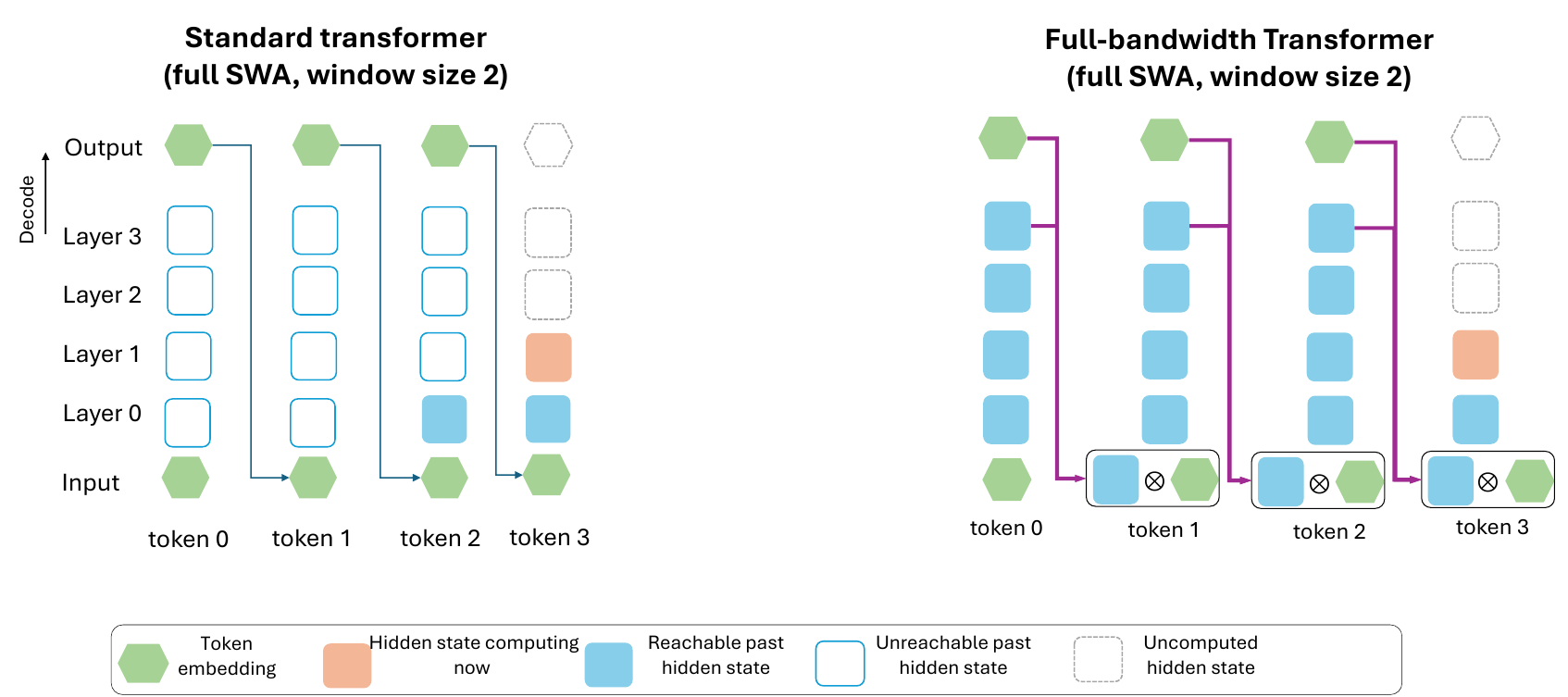}
    \caption{\textbf{Full-bandwidth decoding complements sliding-window attention.}
Consider an idealized SWA-only model; standard decoding can access only hidden states within the local receptive field (left). Full-bandwidth decoding additionally propagates the previous top-layer state through latent feedback (right), providing a recurrent path for information to persist beyond the attention window.
}
    \label{fig:swa_full_bandwidth}
\end{figure}

Sliding-window attention (SWA) reduces the KV-cache and attention cost by restricting each token to a local context, but correspondingly limits direct access to distant hidden states. As illustrated in Fig.~\ref{fig:swa_full_bandwidth} left, under standard decoding a token can only access hidden states that fall within the SWA receptive field. Information from more distant tokens must therefore propagate through successive local-attention layers.

Full-bandwidth decoding introduces an additional recurrent communication path, which enables information that has reached $\bh_{t-1}^{L}$ to continue to propagate even after its originating token falls outside the attention window:
\begin{equation}
    \bh_{t-k}^{L}
    \rightarrow
    \bh_{t-k+1}^{L}
    \rightarrow \cdots \rightarrow
    \bh_{t-1}^{L}
    \rightarrow
    \bu_t.
\end{equation}

Thus, SWA provides high-bandwidth local memory through the KV cache, while latent feedback provides a persistent, compressed channel for information beyond the local window. Importantly, latent feedback does not recover individual hidden states that have left the window; instead, their information can be summarized and carried forward through the recurrent hidden state.


\section{Inference latency}\label{sec:latency}

In the actual vLLM implementation, the extra overhead comes from two parts: The host side for moving around hidden states and the runner side from the extra glu cross gate.

In the end-to-end benchmark under batch size of 1, performed on one H100, latent feedback decoding reduces the throughput by only 2 percent.
\begin{table}[h]
\centering
\begin{tabular}{rrrrrr}
\toprule
 \stds tok/s & \soft tok/s & \soft/\stds & slowdown & \stds ms/step & \soft ms/step \\
\midrule
 214.1 & 209.1 & 0.98 & 2.3\% & 4.670 & 4.782 \\
\bottomrule
\end{tabular}
\caption{Throughput and per-step latency comparison.}
\label{tab:softmax-slowdown}
\end{table}


\newpage
\section{Full-bandwidth transformer carries richer information in shallow-layer residuals}\label{sec:state_tracking}

\begin{figure}[!t]
    \centering
    \begin{subfigure}{\linewidth}
        \includegraphics[width=\linewidth]{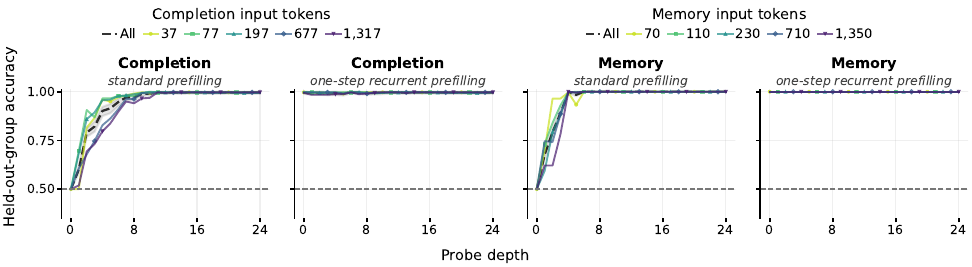}
        \caption{\textbf{State retrieval.} Each sequence specifies either a binary relation between two counters (``Completion'') or a stored absolute binary value (``Memory''),
        followed by varying number (denoted by line color) of label-independent distraction tokens.
        One recurrent step makes the target state nearly perfectly decodable at layer \(0\) across input lengths, whereas standard prefilling requires multiple layers to reconstruct it from the prefix.}
    \end{subfigure}
    \begin{subfigure}{\linewidth}
        \includegraphics[width=\linewidth]{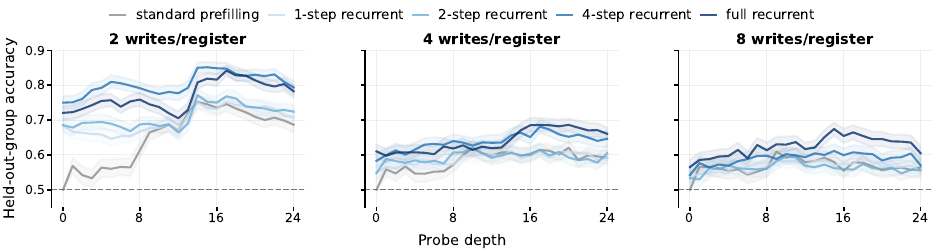}
        \caption{
        \textbf{Multi-register latest-write tracking.} 
        Each sequence performs 2,4, or 8 writes to each of eight binary registers and then queries one register's latest value. We probe its value at each residual depth.
        Recurrent prefilling improves shallow standard accessibility over standard prefilling;
        One recurrent step's gain diminishes in deeper layers and in inputs with more overwriting, where full recurrent performs the best, showing the benefit of maintaining state throughout the sequence.
        }
    \end{subfigure}
    \caption{
    \textbf{Full-bandwidth transformer exposes global state to shallow layers.}
    On three synthetic tasks, we linearly probe the final input token's residual stream across depth (0 denotes input) to predict a binary state of the input (See Appendix~\ref{sec:state_tracking_details} for details);
    We compared standard prefilling, which uses token embedding as inputs, with recurrent prefilling, where the preceding token's top-layer state is fused into the current token's input, similar to Eq.~\eqref{eq:decode-unroll} but uses input token rather than sampled token; \(k\)-step recurrence applies this fusion over the final \(k\) tokens (at the cost of $k+1$ forward pass), while full recurrence applies it throughout the task sequence (at the cost of fully sequential prefill).
    }
    \label{fig:state_tracking}
\end{figure}

To verify the added bandwidth directly, we run controlled state-tracking experiments in which the target is fixed but the intervening context varies (full construction in App.~\ref{sec:state_tracking_details}). Two tasks isolate the effect. \emph{Completion tracking} asks whether a completed counter has reached a required one after a run of no-op updates; \emph{delayed memory} asks the model to recover an initial binary state after a sequence of label-independent scratch operations. Both end at a shared colon, and the label is determined entirely by information before it, so a probe at that colon measures how much of the global state each layer has already reconstructed.

We compare two prefilling regimes. Under \emph{standard prefilling}, the final token enters as its plain embedding; under \emph{one-step recurrent prefilling}, that embedding is fused with the preceding token's top-layer state (Eq.~\eqref{eq:glu_cross}), exactly the layer-0 input latent feedback supplies at decode time. We then fit a linear probe for the target (\textsc{done}/\textsc{more} or \textsc{zero}/\textsc{one}) at each residual-stream depth.

The two regimes differ sharply at the bottom of the stack. Under standard prefilling, a shallow residual can read only the layer-matched, partially processed prefix (the reachability constraint of Eq.~\eqref{eq:std-reach}), so reconstructing the global state takes several layers of further computation; the layer-0 probe is near chance. Recurrent prefilling instead exposes a fully processed prefix summary at the layer-0 input, and layer-0 probe accuracy rises to $99.6\%$ for completion tracking and $100\%$ for delayed memory. Recurrence thus provides a high-bandwidth shortcut that transports globally aggregated information into shallow computation, the mechanism the full-bandwidth view predicts.

One caveat bears emphasis: improved \emph{decodability} does not by itself imply improved \emph{output}. That a target is linearly recoverable at layer 0 shows the information is present, not that the model uses it to decide the next token; making state available and causally exploiting it are distinct, and only the downstream task results (Sec.~\ref{sec:experiments}) speak to the latter.


\section{Explanation on state tracking tasks}\label{sec:state_tracking_details}

We construct paired synthetic examples whose label is determined by information appearing before a shared final colon. The target token itself is never included in the input. We append \(0, 8, 32, 128,\) or \(256\) semantically null scratch updates, allowing us to vary sequence length without changing the target. At the final colon, we record the layer-0 input and the output of every Transformer block.

\paragraph{Completion tracking.}
Each input specifies a required count \(a\) and a completed count \(b\). The target is \textsc{done} if \(a=b\) and \textsc{more} otherwise. For each unordered numeral pair \(\{a,b\}\), we include all four assignments \((a,a),(a,b),(b,a),(b,b)\), balancing every numeral across fields and labels. A representative matched pair, abbreviated to show eight repeated distractors, is

\begin{verbatim}
required = 4                 required = 4
completed = 9                completed = 4
scratch = 7                  scratch = 7
scratch += 0                 scratch += 0
  ... (8 updates)              ... (8 updates)
Status:                      Status:
\end{verbatim}

The left target is \textsc{more}, whereas the right target is \textsc{done}. The two examples share the required count, scratch context, distractor sequence, and final token; only the relation between the two counters changes.

\paragraph{Delayed memory.}
Each input first assigns a binary state and then presents label-independent scratch operations. The target is \textsc{zero} or \textsc{one} according to the initial state. For example,

\begin{verbatim}
state = 0                    state = 1
scratch = 0                  scratch = 0
scratch ^= 0                 scratch ^= 0
scratch ^= 1                 scratch ^= 1
scratch ^= 1                 scratch ^= 1
scratch ^= 0                 scratch ^= 0
scratch ^= 1                 scratch ^= 1
scratch += 0                 scratch += 0
  ... (8 updates)              ... (8 updates)
# final state:               # final state:
\end{verbatim}

The corresponding targets are \textsc{zero} and \textsc{one}. Thus the model must retain the initial bit while processing an identical intervening context. Completion tracking tests a relational state computed from multiple fields, whereas delayed memory tests persistent transport of an already specified state.

\paragraph{Multi-register latest-write tracking.}
We additionally test whether recurrent prefilling can expose several independently updated variables. An input assigns binary values to registers \(r_0,\ldots,r_{m-1}\), performs eight label-independent scratch updates, and then queries one register. The target is \textsc{zero} or \textsc{one} according to that register's most recent assignment. For example, the following matched inputs share the complete update history and differ only in the queried register:

\begin{verbatim}
r4 = 0                      r4 = 0
r4 = 1                      r4 = 1
r0 = 1                      r0 = 1
r7 = 0                      r7 = 0
  ... (10 assignments)        ... (10 assignments)
r7 = 1                      r7 = 1
r1 = 0                      r1 = 0
scratch = 7                 scratch = 7
scratch += 0                scratch += 0
  ... (7 updates)             ... (7 updates)
query = r0                   query = r1
Value:                       Value:
\end{verbatim}

Here the latest values are \(r_0=1\) and \(r_1=0\), so the left target is \textsc{one} and the right target is \textsc{zero}. The model must therefore preserve the latest value of every register and bind the final query to the appropriate component of that state.

\paragraph{Probe construction.}
We train an \(L_2\)-regularized linear classifier at each residual-stream depth using four-fold grouped cross-validation. Completion splits hold out entire unordered numeral-pair groups, and memory splits hold out complete scratch-context groups. The enlarged experiment contains \(1{,}600\) completion examples from 80 groups and \(1{,}280\) memory examples from 128 groups. Because every example ends at the same colon token, the standard layer-0 representation contains no label information beyond the shared token embedding; any above-chance accessibility must be introduced by processing the prefix or by recurrent fusion.

\paragraph{Register-count and overwrite sweeps.}
In the register-count sweep, every input contains 16 assignments and eight null updates and is padded to exactly 180 tokens; only the number of registers varies over \(m\in\{1,2,4,8\}\). This separates the effect of maintaining more variables from input length and total update count. We use 128 structural groups per register count. Each group contains a random register-update schedule, its bitwise value complement, and queries for every register, and grouped cross-validation holds out the entire schedule and all associated queries. The resulting sweep contains \(3{,}840\) examples per prefill condition. To vary overwrite interference directly, we then fix \(m=8\) and use \(2,4,\) or \(8\) writes per register. Each setting contains \(2{,}048\) examples from 128 groups and produces inputs of 180, 276, and 468 tokens, respectively.

\paragraph{Recurrent-suffix controls.}
Besides standard and full recurrent prefilling, we recurrently prefill only the final \(k\in\{1,2,4\}\) input tokens while standard-prefilling the preceding prefix. One step fuses state only at the shared final colon, two steps recurrently process \texttt{Value:}, and four steps additionally include the queried-register digit and newline. We probe the residual stream at the final colon at layers \(0,1,2,\) and \(4\), as well as at every remaining depth, using the same grouped \(L_2\)-regularized classifiers. This sweep distinguishes information accumulated throughout the update sequence from information made accessible locally while processing the final query.

\newpage
\section{Ablation study}\label{sec:ablation_study}

In this section, we perform ablation studies for several of our design choices.

\begin{figure}[!h]
\begin{subfigure}[t]{.45\linewidth}
        \includegraphics[width=\linewidth]{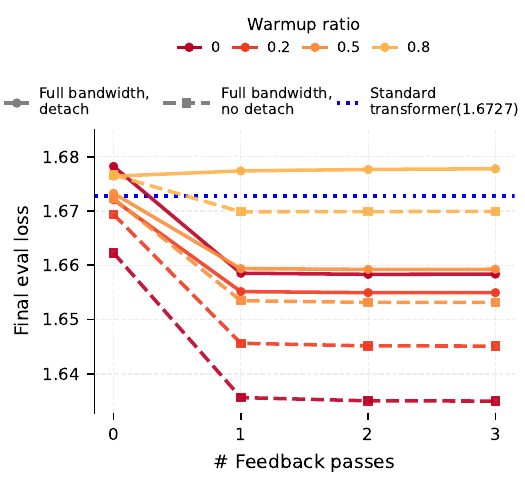}
        \caption{Detaching the hidden state and longer warmup both reduce the gain of feedback pass. Feedback pass 0 denotes standard prefilling with embedding as inputs.}\label{fig:hs_detach}
    \end{subfigure}~~
\begin{subfigure}[t]{.45\linewidth}
        \includegraphics[width=\linewidth]{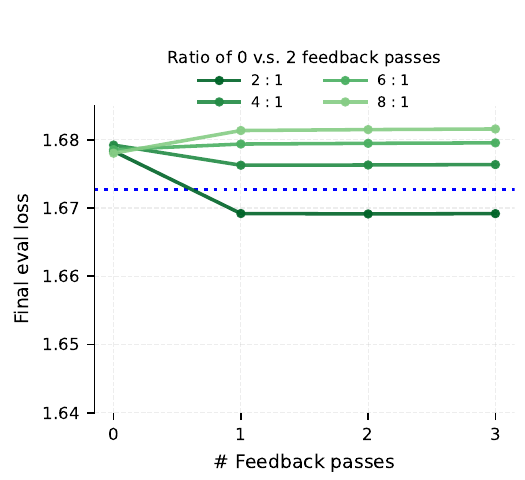}
        \caption{Interleaving between one and multi-pass does not give more efficient training.
        The hope is that such interleaved training can reduce the number of multi-pass steps while preserving the performance.
        }\label{fig:interleaving_schedule}
    \end{subfigure}
\caption{\textbf{Ablation study on feedback scheduling and hidden state detachment.} Training a 1B model on 10B tokens, more iterations with multiple forward passes (darker line, shorter warmup ratio) gives better validation loss, for both first pass and the second pass. Having standard training and multi pass training in bulk provides better performance than interleaving two regimes.}
\end{figure}

\paragraph{Detachment of hidden states} We consider detaching the hidden states after each feedback pass during training, which reduces memory footprint. The solid lines in Fig.~\ref{fig:hs_detach} show the results: Models trained under detachment still show improved validation loss in second pass, however their level of improvement is much smaller compared with the non-detached version (dashed lines.)

\paragraph{Alternative scheduling regime} In the current experiments, we introduced feedback passes later on in the training. Lines of different colors in Fig.~\ref{fig:hs_detach} consider introducing multi forward pass at different stage, where warmup ratio of $0$ denotes multi pass from the beginning.
Overall, the longer the multi pass stage is, the more improvment we get.
We also considered alternative scheduling, where we interleaved between one pass and multi pass at different iterations. 
Fig.~\ref{fig:interleaving_schedule} shows that that it fails to give performance on par with having single and multi pass as two separated stages.
And a bigger issue is that such interleaved training prevents us from starting from an intermediate checkpoint.

\begin{figure}[!h]
    \centering
    \includegraphics[width=\linewidth]{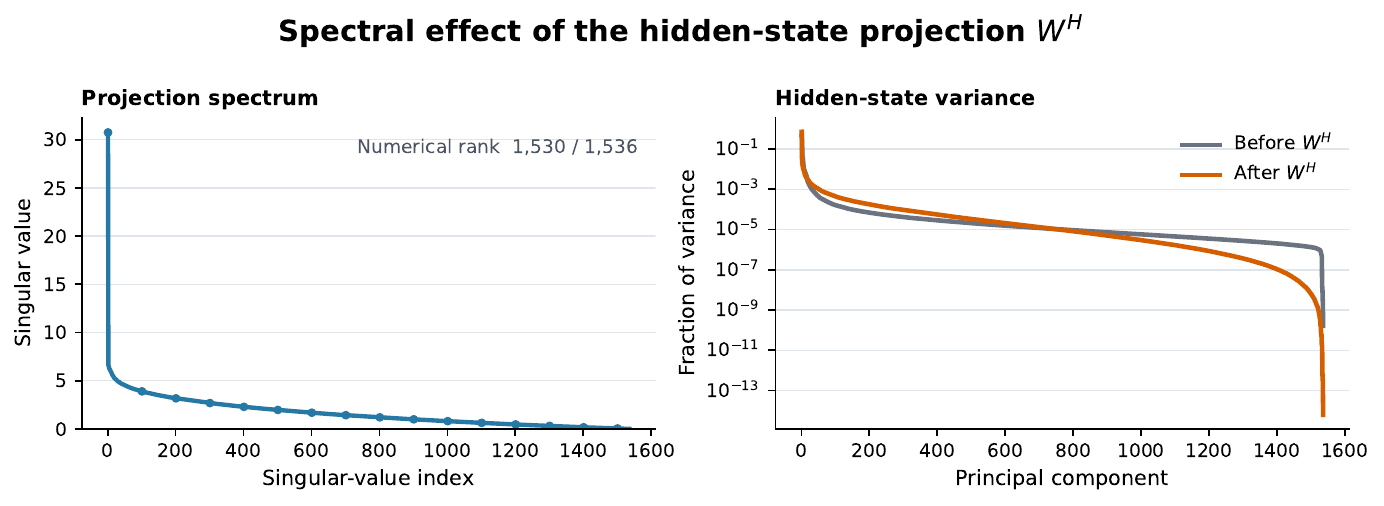}
    \includegraphics[width=\linewidth]{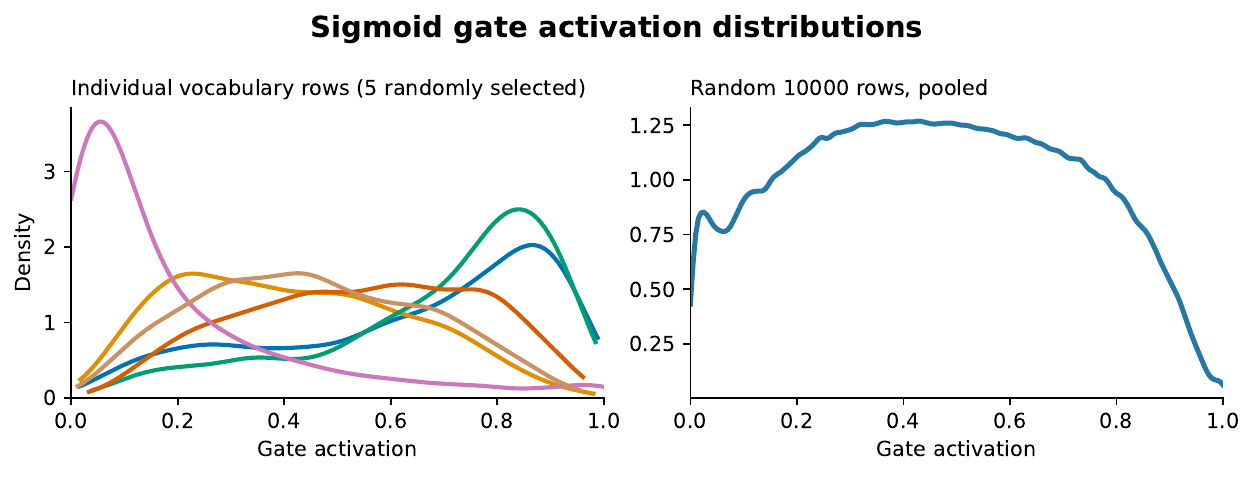}
\caption{
\textbf{Inspection of the fusion gate for a model trained on 200B tokens.}
\textit{Top:} the hidden-state projection $\bWU$ is nearly full rank and largely preserves the variance spectrum of hidden states.
\textit{Bottom:} sigmoid gate activations $\sigma(\bWG\embd_t)$ vary broadly across dimensions and tokens, indicating token-dependent modulation of the recurrent state.
}
    \label{fig:projection_matrix_role}
\end{figure}

\paragraph{Role of projection matrices}
The fusion gate (Eq.~\eqref{eq:glu_cross}) contains two projections, $\bWU$ and $\bWG$, which respectively transform the previous hidden state and the current token embedding. In Fig.~\ref{fig:projection_matrix_role}, we examine whether these projections bottleneck / truncate information or have collapsed outputs.
The hidden-state projection $\bWU$ remains nearly full rank (1530/1536) with a smoothly decaying singular spectrum, and applying it largely preserves the variance spectrum of the observed hidden states, although low-variance directions are somewhat attenuated. Thus, $\bWU$ does not appear to collapse the hidden state into a low-dimensional subspace.
Meanwhile, the sigmoid activations $\sigma(\bWG \embd_t)$ span a broad range between zero and one and vary substantially across tokens, suggesting that $\bWG$ primarily provides token-dependent modulation rather than a fixed dimensional mask. Together, these results suggest that the fusion gate preserves a high-dimensional channel from the previous hidden state while allowing the current token to selectively reweight it.

\begin{figure}[!h]
    \centering
\includegraphics[width=.95\linewidth]{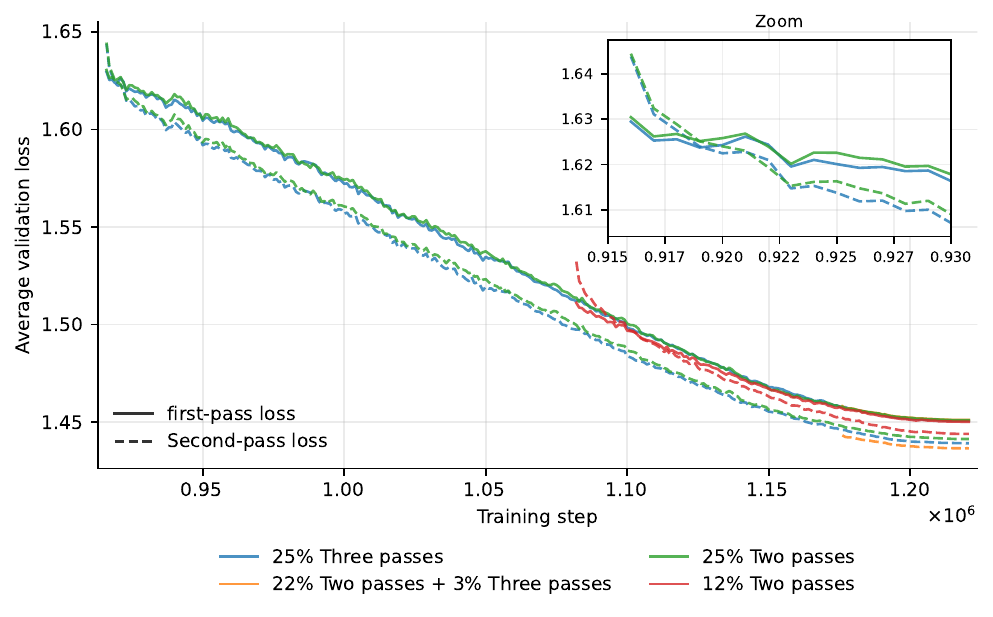}
    \caption{
    Example training dynamics of full-bandwidth transformer for the 400B token run. 
    The feedback pass is only introduced at the cooldown phase (starting from 75\% of the training) 
    }
    \label{fig:training_dynamics}
\end{figure}

\paragraph{Training dynamics} We present examples of what the loss curve looks like during training in Fig.~\ref{fig:training_dynamics}.
The second pass loss (dashed lines) begins above the first pass loss (solid line) and then gradually goes below the first pass.
This pattern is useful for debugging the training: A successful training configuration should result in the dashed lines going under the solid line, which implies that the model learns to read and use its own hidden states for improving the context understanding, a necessary (but insufficient) condition for latent feedback decoding to work.

\begin{figure}[!h]
    \centering
    \includegraphics[width=.95\linewidth]{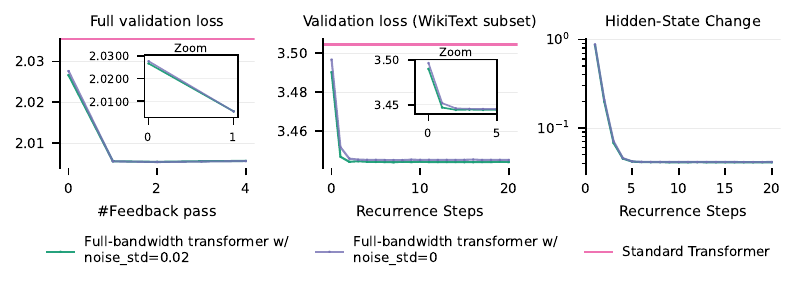}
    \caption{Ablation study on the effect of noise (Eq.~\eqref{eq:jitter_noise}). Evaluated on 1B scale run, jitter noise improves the loss, especially on the first pass loss.} 
    \label{fig:jitter_noise_ablate}
\end{figure}

\paragraph{Role of jitter noise} Lastly, we ablate the role of the jitter noise (Eq.~\eqref{eq:jitter_noise}). Fig.~\ref{fig:jitter_noise_ablate} shows the loss v.s. feedback passes on two models trained on 1B tokens, with and without jitter noise, where we observe that jitter noise slightly improves the loss, especially under standard prefilling. The jitter noise is not observed to change the stable attractor pattern.

\subsection{Symmetric gate}\label{app:symmetric_gate}
\begin{figure}
    \centering
    \includegraphics[width=\linewidth]{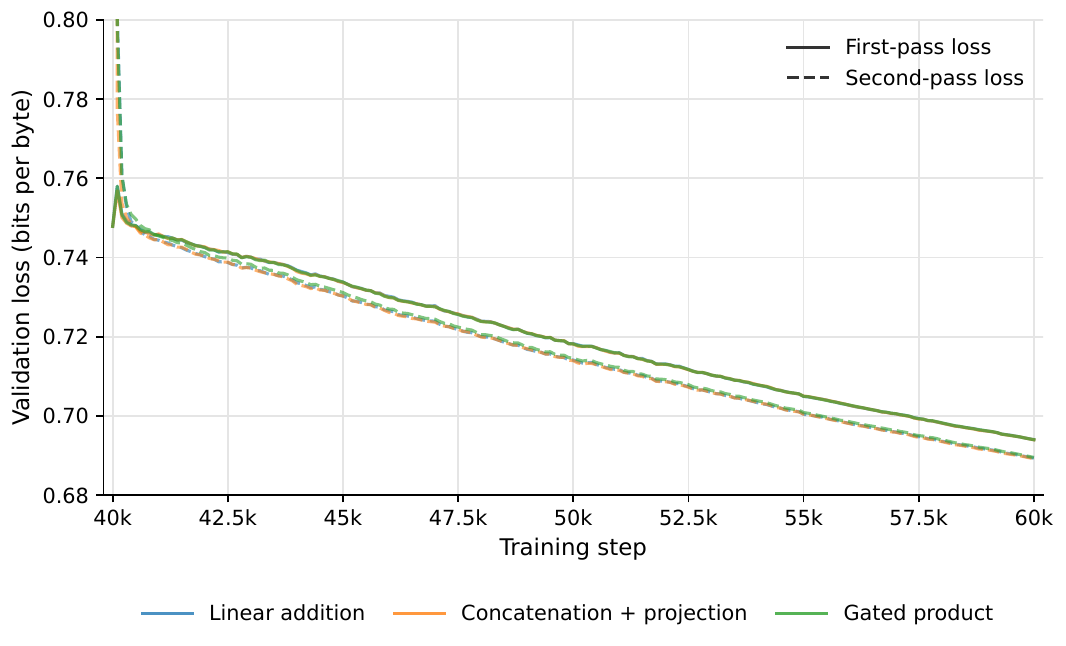}
    \caption{Tested on a small-scale Nanochat run, fused inputs under symmetric gate still manage to improve the validation loss.
    }
    \label{fig:fusion_gate_ablation}
\end{figure}

Evaluated on a separate smaller-scale setting: A 20-layer Nanochat model (950M parameters) trained on 20B tokens using a modified Nanochat codebase. We consider three types of fusion gates
\begin{itemize}
    \item Gated product: Same as the glu cross fusion (Eq.~\eqref{eq:glu_cross}) used in the main text.
    \item Linear addition: $\bWU \bh_{t-1} + \bWG \be_t$.
    \item Concatenation and projection:$\bWR [\bh_{t-1}; \be_t],
    \qquad \bWR \in \mathbb{R}^{D \times 2D}$.
\end{itemize}
The results are shown in Fig.~\ref{fig:fusion_gate_ablation}, where we do not see one method outperform another one on a small scale, nor do we observe that symmetric fusion shows signs of shortcut seeking.
This may suggest that future large-scale experiments can consider alternative forms of fusion.

\newpage
\section{Model output}\label{sec:extra_model_outputs}

\begin{figure}[!h]
\begin{tcolorbox}[colback=white, colframe=black!60,
    title=\textbf{Problem}]
\small
What is the last nonzero digit to the right of the decimal point in the decimal
expansion of $\frac{137}{500}$?
\end{tcolorbox}

\begin{tcolorbox}[colback=gray!5, colframe=gray!50,
    title=\textbf{Standard decoding} (200B model)]
\small
To determine the last nonzero digit to the right of the decimal point in the
decimal expansion of $\frac{137}{500}$, we start by performing the division of
137 by 500.

First, we set up the division:
\[ 137 \div 500 \]
Since 137 is less than 500, the decimal expansion will start with a zero. {\color{blue}\textit{[...326 tokens omitted...]}}

Since the remainder is now $0$, the decimal expansion of $\frac{137}{500}$ is:
\[ 0.274 \]
The last nonzero digit to the right of the decimal point is $\boxed{4}$.
\end{tcolorbox}

\begin{tcolorbox}[colback=blue!5, colframe=blue!40,
    title=\textbf{Soft decoding} (200B model)]
\small
To find the last nonzero digit to the right of the decimal point in the decimal
expansion of $\frac{137}{500}$, we can perform the division and observe the
pattern of the decimal expansion.

$\frac{137}{500} = 0.274$

The last nonzero digit to the right of the decimal point in the decimal
expansion of $\frac{137}{500}$ is 4.
\end{tcolorbox}
\caption{Qualitative comparison of outputs from the 200B model under standard
and soft decoding (greedy decoding).
Both reach the correct answer; soft decoding is substantially
more concise. Truncated text is indicated by [...].
This is no longer observable on instruction-tuned version.
}
\label{fig:qualitative-decimal}
\end{figure}

\begin{tcolorbox}[colback=blue!5, colframe=blue!40,
    title=\textbf{Soft decoding}]
\small
**** 200b model, standard decoding ****

 To determine the last nonzero digit to the right of the decimal point in the decimal expansion of \(\frac{137}{500}\), we start by performing the division of 137 by 500.

First, we set up the division:
\[
137 \div 500
\]

Since 137 is less than 500, the decimal expansion will start with a zero. We can rewrite the division as:
\[
137.0 \div 500
\]

Next, we perform the division step-by-step:
1. \(500\) goes into \(1340\) (the first two digits of \(137.0\)) twice:
   \[
   500 \times 2 = 1000
   \]
   Subtract \(1000\) from \(1340\):
   \[
   1340 - 1000 = 340
   \]

2. Bring down the next digit (0), making it \(340.0\):
   \[
   500 \times 0 = 0
   \]
   Subtract \(0\) from \(340\):
   \[
   340 - 0 = 340
   \]

3. Bring down the next digit (0), making it \(3400\):
   \[
   500 \times 6 = 3000
   \]
   Subtract \(3000\) from \(3400\):
   \[
   3400 - 3000 = 400
   \]

4. Bring down the next digit (0), making it \(4000\):
   \[
   500 \times 8 = 4000
   \]
   Subtract \(4000\) from \(4000\):
   \[
   4000 - 4000 = 0
   \]

Since the remainder is now \(0\), the decimal expansion of \(\frac{137}{500}\) is:
\[
0.274
\]

The last nonzero digit to the right of the decimal point is:
\[
\boxed{4}
\]

------------------------------

**** 200b model, soft ****

 To find the last nonzero digit to the right of the decimal point in the decimal expansion of $\frac{137}{500}$, we can perform the division and observe the pattern of the decimal expansion.

$\frac{137}{500} = 0.274$

The last nonzero digit to the right of the decimal point in the decimal expansion of $\frac{137}{500}$ is 4.
\end{tcolorbox}

\begin{tcolorbox}[colback=blue!5, colframe=blue!40,
    title=\textbf{Soft decoding}]
\small
**** 200b model, standard decoding ****

 To simplify the expression \(\frac{(10r^3)(4r^6)}{8r^4}\), we will follow these steps:

1. **Multiply the numerator**:
   \[
   (10r^3)(4r^6)
   \]
   Using the properties of exponents, we can combine the terms:
   \[
   10 \cdot 4 \cdot r^{3+6} = 40r^9
   \]

2. **Rewrite the expression with the simplified numerator**:
   \[
   \frac{40r^9}{8r^4}
   \]

3. **Simplify the fraction**:
   - First, simplify the coefficients:
     \[
     \frac{40}{8} = 5
     \]
   - Next, simplify the exponents of \(r\):
     \[
     \frac{r^9}{r^4} = r^{9-4} = r^5
     \]

4. **Combine the simplified parts**:
   \[
   5r^5
   \]

Thus, the simplified form of the expression \(\frac{(10r^3)(4r^6)}{8r^4}\) is:
\[
\boxed{5r^5}
\]
------------------------------

**** 200b model, soft ****

 To simplify the expression, we can first simplify the numerator and then divide by the denominator.

Numerator: $(10r^3)(4r^6) = 40r^9$

Denominator: $8r^4$

Now, we divide the numerator by the denominator:

$\frac{40r^9}{8r^4} = 5r^{9-4} = 5r^5$

So, the simplified expression is $5r^5$.
\end{tcolorbox}


\end{document}